\documentclass[sigconf]{acmart}
\usepackage[english]{babel}
\newtheorem{theorem}{Theorem}
\usepackage[ruled,linesnumbered]{algorithm2e}
\newtheorem{lemma}[theorem]{Lemma}
\usepackage{makecell}
\usepackage{algpseudocode}

\usepackage{hyperref}
\usepackage{graphicx, subcaption}
\usepackage{enumitem}
\usepackage{multirow}
\usepackage{verbatim}
\usepackage[ruled,linesnumbered]{algorithm2e}
\definecolor{syd_color}{rgb}{0, 0, 0}
\newcommand{\syd}[1]{\textcolor{syd_color}{#1}}
\newcommand{\ie}{\emph{i.e., }}
\newcommand{\eg}{\emph{e.g., }}
\newcommand{\etal}{\emph{et al.}}

\newcommand{\wrt}{\emph{w.r.t. }}

\newcommand{\Lapl}{\mathbf{\mathop{\mathcal{L}}}}

\newcommand{\Space}[1]{\mathbb{#1}}
\newcommand{\Mat}[1]{\mathbf{#1}}
\newcommand{\Set}[1]{\mathcal{#1}}

\AtBeginDocument{%
  \providecommand\BibTeX{{%
    \normalfont B\kern-0.5em{\scshape i\kern-0.25em b}\kern-0.8em\TeX}}}

\copyrightyear{2023}
\acmYear{2023}
\setcopyright{acmlicensed}\acmConference[WWW '23]{Proceedings of the ACM Web Conference 2023}{April 30-May 4, 2023}{Austin, TX, USA}
\acmBooktitle{Proceedings of the ACM Web Conference 2023 (WWW '23), April 30-May 4, 2023, Austin, TX, USA}
\acmPrice{15.00}
\acmDOI{10.1145/3543507.3583521}
\acmISBN{978-1-4503-9416-1/23/04}

\begin{document}

\title{GIF: A General Graph Unlearning Strategy via Influence Function }

\author{Jiancan Wu$^1$\footnotemark[1], Yi Yang$^1$\footnotemark[1], Yuchun Qian$^1$, Yongduo Sui$^1$, Xiang Wang$^{1}$\footnotemark[2], Xiangnan He$^{1}$\footnotemark[2]}
\affiliation{
\institution{$^1$University of Science and Technology of China \country{China},
}}
\email{wujcan@gmail.com, {yanggnay,daytoy_qyc,syd2019}@mail.ustc.edu.cn,{xiangwang1223,xiangnanhe}@gmail.com}

% \author{Jiancan Wu\footnotemark[1]}
% \email{wujcan@gmail.com}
% \orcid{0000-0002-6941-5218}
% \affiliation{%
%   \institution{University of Science and Technology of China}
%   \city{Hefei}
%   \state{Anhui}
%   \country{China}
%   \postcode{230026}
% }

% \author{Yi Yang\footnotemark[1]}
% \email{yanggnay@mail.ustc.edu.cn}
% \orcid{0000-0002-5687-7819}
% \affiliation{%
%   \institution{University of Science and Technology of China}
%   \city{Hefei}
%   \state{Anhui}
%   \country{China}
%   \postcode{230026}
% }

% \author{Yuchun Qian}
% \email{daytoy_qyc@mail.ustc.edu.cn}
% \orcid{0000-0003-4135-6823}
% \affiliation{%
%   \institution{University of Science and Technology of China}
%   \city{Hefei}
%   \state{Anhui}
%   \country{China}
%   \postcode{230026}
% }

% \author{Yongduo Sui}
% \email{syd2019@mail.ustc.edu.cn}
% \orcid{0000-0003-4492-147X}
% \affiliation{%
%   \institution{University of Science and Technology of China}
%   \city{Hefei}
%   \state{Anhui}
%   \country{China}
%   \postcode{230026}
% }

% \author{Xiang Wang\footnotemark[2]}
% \email{xiangwang1223@gmail.com}
% \orcid{0000-0003-1317-9567}
% \affiliation{%
%   \institution{University of Science and Technology of China}
%   \city{Hefei}
%   \state{Anhui}
%   \country{China}
%   \postcode{230026}
% }

% \author{Xiangnan He\footnotemark[2]}
% \email{xiangnanhe@gmail.com}
% \orcid{0000-0001-7231-594X}
% \affiliation{%
%   \institution{University of Science and Technology of China}
%   \city{Hefei}
%   \state{Anhui}
%   \country{China}
%   \postcode{230026}
% }

\def \authors{Jiancan Wu, Yi Yang, Yuchun Qian, Yongduo Sui, Xiang Wang, Xiangnan He}

\renewcommand{\shortauthors}{Jiancan Wu, et al.}
\begin{abstract}
% With the overall societal emphasis on privacy and security, everyone has the right to request a
% trained GNN model for unlearning data related to themselves. However, machine unlearning methods can not be directly applied to the domain of graph unlearning due to the dependencies between graph-structured data and current graph unlearning approaches are either limited to linear GNN models or hard to achieve satisfying performance-complexity trade-offs.
% Considering the characteristics of graph topology, , we propose a more general and powerful graph unlearning approximation algorithm GIF, which can conveniently process graph unlearning tasks with the help of mature auto-grad systems and estimation methods. By analyzing the actual meaning of each item of GIF on Simple Graph Convolution, it is enlightening to understand the black box of graph unlearning process. Experimental shows that on major benchmarks show GIF show excellent model performance and efficient estimation. 
With the greater emphasis on privacy and security in our society, the problem of graph unlearning --- revoking the influence of specific data on the trained GNN model, is drawing increasing attention. However, ranging from machine unlearning to recently emerged graph unlearning methods, existing efforts either resort to retraining paradigm, or perform approximate erasure that fails to consider the inter-dependency between connected neighbors or imposes constraints on GNN structure, therefore hard to achieve satisfying performance-complexity trade-offs. 

In this work, we explore the influence function tailored for graph unlearning, so as to improve the unlearning efficacy and efficiency for graph unlearning.
We first present a unified problem formulation of diverse graph unlearning tasks \wrt node, edge, and feature.
Then, we recognize the crux to the inability of traditional influence function for graph unlearning, and devise Graph Influence Function (GIF), a model-agnostic unlearning method that can efficiently and accurately estimate parameter changes in response to a $\epsilon$-mass perturbation in deleted data. 
The idea is to supplement the objective of the traditional influence function with an additional loss term of the influenced neighbors due to the structural dependency.
Further deductions on the closed-form solution of parameter changes provide a better understanding of the unlearning mechanism.
% Considering the characteristics of graph topology, we propose a more general and powerful graph unlearning approximation algorithm GIF, short for Graph-oriented Influence Function, which can conveniently process graph unlearning tasks with the help of Influence Function.
%% TODO: a sentence about how GIF works.
%% TODO: a sentence about explainability
% By analyzing the actual meaning of each item of GIF on Simple Graph Convolution, it is enlightening to understand the black box of graph unlearning process.
We conduct extensive experiments on four representative GNN models and three benchmark datasets to justify the superiority
of GIF for diverse graph unlearning tasks in terms of unlearning
efficacy, model utility, and unlearning efficiency. Our implementations are available at \url{https://github.com/wujcan/GIF-torch/}.
\end{abstract}


%CCS CONCEPTS 部分需要补足
%\ccsdesc[500]{Computer systems organization~Embedded systems}




\maketitle

\renewcommand{\thefootnote}{\fnsymbol{footnote}} %将脚注符号设置为fnsymbol类型，即特殊符号表示
\footnotetext[1]{These authors contributed equally to this work.} %对应脚注[1]
\footnotetext[2]{Corresponding authors.}

% !TeX root = ./0_main.tex
\section{Introduction}
Machine unlearning \cite{Cao&Yang2015} is attracting increasing attention in both academia and industry.
At its core is removing the influence of target data from the deployed model, as if it did not exist.
It is of great need in many critical scenarios, such as (1) the enforcement of laws concerning data protection or user's right to be forgotten \cite{regulation2018general,pardau2018california,KwakLPL17Let}, and (2) the demands for system provider to revoke the negative effect of backdoor poisoned data \cite{ZhangCHLZFHY22Poison,RubinsteinNHJLRTT09ANTIDOTE}, wrongly annotated data \cite{PangZQJ21Recorrupted}, or out-of-date data \cite{WangLF0LC22Causal}.
As an emerging topic, machine unlearning in the graph field, termed \textbf{graph unlearning}, remains largely unexplored, but is the focus of our work.
Specifically, graph unlearning focuses on ruling out the influence of graph-structured data (\eg nodes, edges, and their features) from the graph neural networks (GNNs) \cite{Kipf2017GCN,GAT,SGC,lightgcn,WuWF0CLX21,LiWZW0C22,DIR}.
Clearly, the nature of GNNs --- the entanglement between the graph structure and model architecture --- is the main obstacle of unlearning.
For example, given a node of a graph as the unlearning target, we need to not only remove its own influence, but also offset its underlying effect on neighbors multi-hop away.

\begin{figure}[t]
    \centering
    \includegraphics[width=0.46\textwidth]{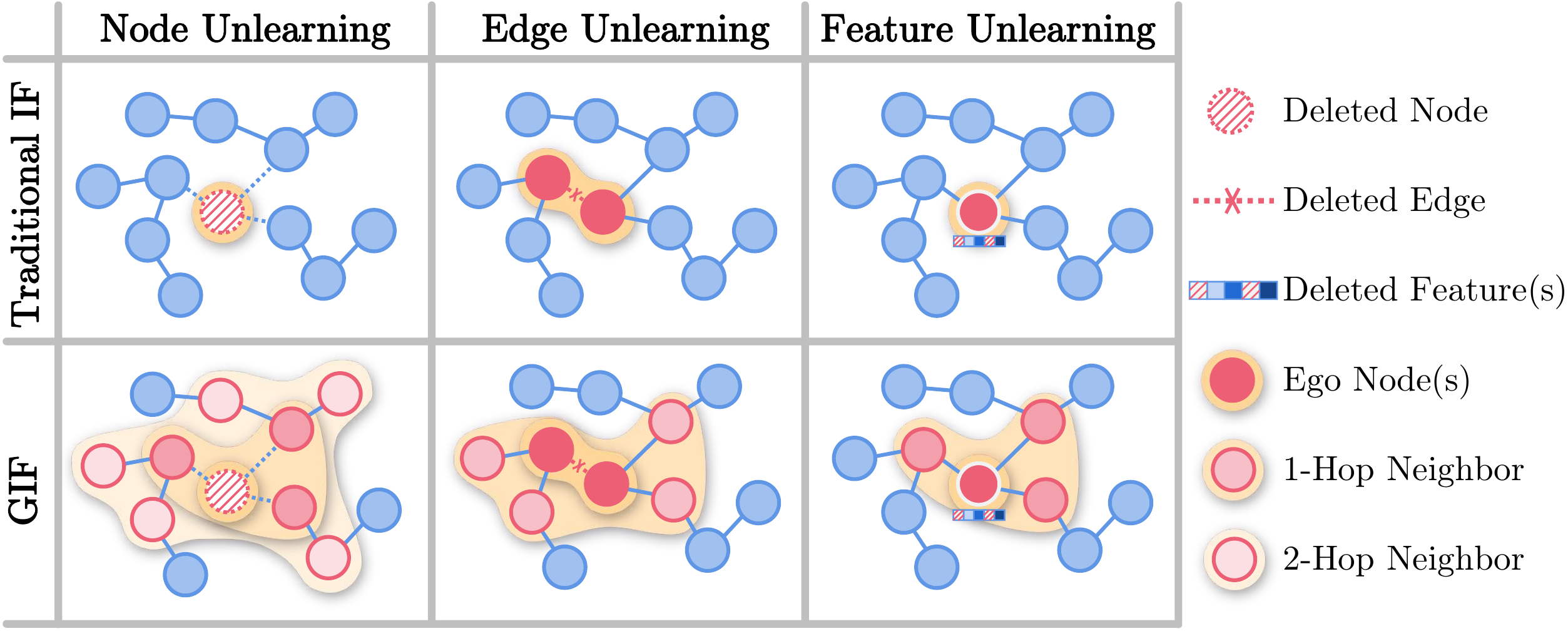}
    \caption{The differences between the traditional influence function method and our GIF. The traditional IF only computes the parameter change if the loss of nodes that are directly affected by the unlearning request (as shown in the colored region) is upweighted by a small infinitesimal amount, while our GIF considers both the directly affected node(s) and the influenced neighborhoods as shown in the bottom subfigures, hence, it can unlearn the data more completely.}
    \Description[The differences between the traditional influence function method and our GIF]{The conventional IF algorithm only calculates the parameter change if the loss of nodes directly impacted by the unlearning request (as indicated in the colored region) is upweighted by a small infinitesimal amount. In contrast, our GIF approach takes into account not only the directly impacted node(s) but also their surrounding neighborhoods, as illustrated in the bottom subfigures. Therefore, it can more thoroughly unlearn the data.}
    \label{fig:method_comparison}
\end{figure}

Such a structural influence makes the current approaches fall short in graph unlearning.
Next we elaborate on the inherent limitations of these approaches:
\begin{itemize}[leftmargin=*]
    \item A straightforward unlearning solution is retraining the model from scratch, which only uses the remaining data. However, it can be resource-consuming when facing large-scale graphs like social networks and financial transaction networks.
    
    \item Considering the unlearning efficiency, some follow-on studies \cite{machine,shardunlearn2,Chen2022Graph,Chen2022recommendation,Cong2022GRAPHEDITOR} first divide the whole training data into multiple disjoint shards, and then train one sub-model for each shard.
    When the unlearning request arrives, only the sub-model of shards that comprise the unlearned data needs to be retrained.
    Then, aggregating the predictions from all the sub-models can get the final prediction.
    Although this exact way guarantees the removal of all information associated with the unlearned data, splitting data into shards will inevitably destroy the connections in samples, especially for graph data, hence hurting the model performance.
    GraphEditor~\cite{Cong2022GRAPHEDITOR} is a very recent work that supports exact graph unlearning free from shard model retraining, but is restricted to linear GNN structure under Ridge regression formulation.
    
    \item Another line \cite{fisher,certified,graphcerti,approxi1,approxigrad1,approxigrad2} resorts to gradient analysis techniques instead to approximate the unlearning process, so as to avoid retraining sub-models from scratch. Among them, influence function \cite{Koh2017Understanding} is a promising proxy to estimate the parameter changes caused by a sample removal, which is on up-weighting the individual loss \wrt the target sample, and then reduces its influence accordingly.
    However, it only considers the individual sample, leaving the effect of sample interaction and coalition untouched. As a result, it hardly approaches the structural influence of graph, thus generalizing poorly on graph unlearning.
    To the best of our knowledge, no effort has been made to tailor the influence function for graph unlearning.
\end{itemize}

To fill this research gap, we explore the graph-oriented influence function in this work.
Specifically, we first present a unified problem formulation of graph unlearning \wrt three essential gradients: node, edge, and feature, thus framing the corresponding tasks of node unlearning, edge unlearning, and feature unlearning. 
Then, inspecting the conventional influence function, we reveal why it incurs incomplete data removal for GNN models.
Taking unlearning on an ego node as an example, it will not only affect the prediction of the node, but also exert influence on the $K$-hop neighboring nodes, due to the message passing scheme of GNN.
Similar deficiencies are observed for edge and feature unlearning tasks.
Realizing this, we devise a \underline{G}raph \underline{I}nfluence \underline{F}unction (\textbf{GIF}) to consider such structural influence of node/edge/feature on its neighbors.
It performs the accurate unlearning answers of simple graph convolutions, and sheds new light on how different unlearning tasks for various GNNs influence the model performance.

We summarize our contributions as follows,
\begin{enumerate}[leftmargin=*]
    \item We propose a more general and powerful graph unlearning algorithm \textbf{GIF} tailor-made for GNN models, which allows us to estimate the model prediction in advance efficiently and protect privacy.
    
    \item We propose more comprehensive evaluation criteria and major tasks for graph unlearning. Extensive experiments show the effectiveness of GIF.
    
    \item To our best knowledge, our GIF is one of the first attempts to interpret the black box of graph unlearning process.
\end{enumerate}

% \input{1_intro}
% \input{1_intorduction}

% \input{2_related}
% !TeX root = ./0_main.tex
\section{Preliminary}

Throughout this paper, we define the lower-case letters in bold (\eg $\Mat{x}$) as vectors. 
The blackboard bold typefaces (\eg $\Space{R}$) denote the spaces, while the calligraphic font for uppercase letters (\eg $\Set{V}$) denote the sets. The notions frequently used in this paper are summarized in Table~\ref{tab:notations}.

\begin{table}
  \caption{Summary of Notations}
  \vspace{-5pt}
  \label{tab:notations}
  \begin{tabular}{ll}
    \toprule
    
    Notations & Description\\
    \midrule
    \syd{$\Set{G}=\{\Set{V},\Set{E},\Set{X}\}$} & Input graph \\
    $\Set{D}\syd{_0} = \{z_{1},z_{2},...,z_{k}\}$ & Training set\\
    $f_{\Set{G}}, f_{\Set{G}\setminus \Delta \Set{G}}$ & 
    % GNN trained on graph $\Set{G}$ and $\Set{G}\setminus \Delta \Set{G}$ from scratch
    \makecell[l]{GNN trained on graph $\Set{G}$ and \\ $\Set{G}\setminus \Delta \Set{G}$ from scratch}\\
    $z_{i}$, $z_{test}$  & \syd{Training} and test node \\
    $f_{\Set{G}}(z_{i};\Set{G}\setminus \Delta \Set{G})$   &   Prediction of $z_i \in \Set{G}\setminus \Delta \Set{G}$ using $f_{\Set{G}}$ \\
    $ l\left(f_{\Set{G}}(z_i),y_i\right) $ &Original loss\\
    $ \Delta l\left(f(z_i),y_i\right)$&Influenced loss\\
    $ l\left(\hat{f}_{\Set{G}}(z_{i}),y_{i}\right)$ & Final loss\\
  \bottomrule
\end{tabular}
\end{table}

\subsection{Graph Unlearning Formulation}
In this work, we focus on the problem of node classification. Consider a graph \syd{$\Set{G}=\{\Set{V},\Set{E},\Set{X}\}$} with $|\Set{V}|$ nodes and $|\Set{E}|$ edges. Each node $v_{i} \in \Set{V}$ is associated with an $F$-dimensional feature vector $\Mat{x}_{i} \in \Set{X}$. The training set $\Set{D}_{0}$ contains $N$ samples $\{z_{1},z_{2},...,z_{N}\}$, each of which is annotated with a label $y \in \Set{Y}$. 
\syd{Given a graph $\Set{G}$,} the goal of node classification task is to train a GNN model that predicts the label of a node $v \in \Set{V}$.
% The goal of node classification task is to train a GNN model that predicts the label of a node $v \in \Set{V}$ given the graph $\Set{G}$ and the node's feature $\Mat{x}_i$.

After the arrival of unlearning request $\Delta \Set{G} = \{ \Set{V}^{(rm)}, \Set{E}^{(rm)}, \Set{X}^{(rm)} \}$, the goal of graph unlearning is to find a mechanism $\mathcal{M}$ that \syd{takes $\Set{D}_0, f_{\Set{G}}$ and $\Delta \Set{G}$ as input}, then outputs a new model $\hat{f}$ parameterized by $\hat{\theta}$ that minimizes the discrepancy $D(\hat{f}, f_{\Set{G}\setminus \Delta \Set{G}})$, where $f_{\Set{G}\setminus \Delta \Set{G}}$ is the GNN model retrained from scratch on the remaining graph $\Set{G}\setminus \Delta \Set{G}$ using the objective similar to Equation~\eqref{eq:task_graph_learning}. The graph unlearning mechanism should satisfy the following criteria:

% After the arrival of unlearning request $\Delta \Set{G} = \{ \Set{V}^{(rm)}, \Set{E}^{(rm)}, \Set{X}^{(rm)} \}$, the goal of graph unlearning is to find a mechanism $\mathcal{M}$ that takes as input $\Set{D}_0, f_{\Set{G}}$ and $\Delta \Set{G}$, then outputs a new model $\hat{f}$ parameterized by $\hat{\theta}$ that minimizes the discrepancy $D(\hat{f}, f_{\Set{G}\setminus \Delta \Set{G}})$, where $f_{\Set{G}\setminus \Delta \Set{G}}$ is the GNN model retrained from scratch on the remaining graph $\Set{G}\setminus \Delta \Set{G}$ using the objective similar to Equation~\eqref{eq:task_graph_learning}. The graph unlearning mechanism should satisfy the following criteria:

\begin{itemize}[leftmargin=*]
        
    \item \textbf{Removal Guarantee}. The primary need of unlearning mechanism is to remove the information of deleted data completely from the trained model, including the deleted data itself and its influences on other samples.
    
    \item \textbf{Comparable Model Utility}. It is a practical requirement that the unlearning mechanism only brings in a small utility gap in comparison to retraining from scratch.
    
    \item \textbf{Reduced Unlearning Time}. The unlearning mechanism should be time-efficient as compared to model retraining.

    \item \textbf{Model Agnostic}. The strategy for unlearning can be applied to any GNN model with \syd{various architectures}, including but not limited to linear GNNs \syd{or} non-linear GNNs.
\end{itemize}
Furthermore, we can categorize the task of graph unlearning into fine-grained tasks based on the type of request $\Delta \Set{G}$. In this work, we consider the following three types of graph unlearning tasks, while leaving the others in future work:
\begin{itemize}[leftmargin=*]
    \item \textbf{Node Unlearning}: $\Delta \Set{G} = \{ \Set{V}^{(rm)}, \varnothing, \varnothing  \}$.
    \item \textbf{Edge Unlearning}: $\Delta \Set{G} = \{ \varnothing, \Set{E}^{(rm)}, \varnothing  \}$.
    \item \textbf{Feature Unlearning}: $\Delta \Set{G} = \{ \varnothing, \varnothing, \Set{X}^{(rm)} \}$.
\end{itemize}

\subsection{Abstract Paradigm of GNNs}

The core of GNNs is to apply the neighborhood aggregation on $\Set{G}$, recursively integrate the vectorized information from neighboring nodes, and update the representations of ego nodes.
Thereafter, a classifier is used to generate the final predictions.

% The core of graph neural network is to apply the neighborhood aggregation on $\Set{G}$, recursively integrate the vectorized information from neighboring nodes, and update the representations of ego nodes.
% Thereafter, a classifier is used to generate the final prediction.

% \begin{align}
%     \Mat{Z} = h(\Set{G}),
% \end{align}
% where $h(\cdot)$ the GNN function to encode connectivity information into representation learning.

\vspace{5pt}\noindent\textbf{Neighborhood Aggregation}. The aggregation scheme consists of \syd{the following} two crucial components:

\noindent(1) Representation aggregation layers. After $k$ layers, a node’s representation is able to capture the structural information within its \syd{$k$-hop} neighbors. The $k$-th layer is formulated as:
\begin{gather}
    \Mat{a}_i^{(k)} = f_{\rm aggregate} \left( {\Mat{z}_j^{(k-1)} | j \in \Set{N}_i} \right)\nonumber,\\
    \Mat{z}_i^{(k)} = f_{\rm combine} \left( \Mat{z}_i^{(k-1)}, \Mat{a}_i^{(k)} \right),
\end{gather}
where $\Mat{a}_i^{(k)}$ denotes the aggregation of vectorized information from node $i$'s neighborhood $\Set{N}_i$ and $\Mat{z}_i^{(k)}$is the representation of node $i$ after $k$ aggregation layers, which integrates $\Mat{a}_i^{(k)}$ with its previous representation $\Mat{z}_i^{(k-1)}$. Specially, $\Mat{z}_i^{(0)} = \Mat{x}_i$.
The designs for aggregation function $ f_{\rm aggregate}(\cdot) $ and combination function $ f_{\rm combine}(\cdot) $ vary in different studies~\cite{Kipf2017GCN,GAT,GIN,SAGE}.

\noindent(2) Readout layer. Having obtained the representations at different layers, the readout function generates the final representations for each node:
\begin{align}
    \Mat{z}_i = f_{\rm readout} \left( \{ \Mat{z}_i^{(k)} | k = [0, \cdots, K \} \right),
\end{align}
which can be simply set as the last-layer representation~\cite{BergKW17GCMC,YingHCEHL18Graph}, concatenation~\cite{Wang0WFC19NGCF}, or weighted summation~\cite{lightgcn} over the representations of all layers.

\vspace{5pt}\noindent\textbf{Prediction and Optimization}. After that, a classifier (or prediction layer) is built upon the final representations of nodes to predict their classes. A classical solution is an MLP network with \textit{Softmax} activation,
\begin{align}
    \hat{y}_i = f_{\Set{G}}(z_i) = \text{Softmax}\left(\Mat{z}_i W^{(pred)}\right),
\end{align}
where $f_{\Set{G}}$ is \syd{a} GNN model built on $\Set{G}$ parameterized by $\theta_{0}$, $W^{(pred)} \in \theta_{0}$ is the parameters of the classifier, $\hat{y}_i$ is the predicted probability of sample $z_i$.
The optimal model parameter $\theta_{0}$ is obtained by  minimizing the following objective:
\begin{align}\label{eq:task_graph_learning}
    \theta_0 = \mathop{\arg\min}_{\theta} \Lapl_0, \quad \Lapl_0 = \sum_{z_i \in \Set{D}_0} l\left(f_{\Set{G}}(z_i), y_i\right),
\end{align}
where $l$ is the loss function defined on the prediction of each sample $f_{\Set{G}}(z_i)$ and its corresponding label $y_i$, $\Lapl_0$ is the total loss for all training samples.

\subsection{Traditional Influence Functions}
The concept of influence function comes from robust statistics, which measures how the model parameters change when we upweight a sample by an infinitesimally-small amount.

\begin{proposition} 
 (Traditional Influence Functions): Assuming we get a node unlearning request $\Delta \Set{G} = \{ \Set{V}^{(rm)}, \varnothing, \varnothing  \}$ for the deployed model $f_{\Set{G}}$ learned under Equation~\eqref{eq:task_graph_learning}, then the estimated parameter change using traditional influence function is $\hat{\theta} - \theta_{0} \approx   H_{\theta_{0}}^{-1} \nabla_{\theta_{0}} \Lapl_{\Delta \Set{G}}$, where $\theta_{0}$ and $\hat{\theta}$ are the model parameters of the learned model before and after unlearning, respectively, $H_{\theta_{0}}$ is the Hessian matrix of $\Lapl_0$ \wrt $\theta_0$, $\Lapl_{\Delta \Set{G}} = \sum_{z_i \in \Delta \Set{D}}l\left(f_{\Set{G}}(z_i), y_i\right)$
\end{proposition}

\renewcommand{\thefootnote}{\arabic{footnote}}

Following the traditional influence function algorithms \cite{influencesta}, the loss defined on original training set for node classification tasks can be described as
\begin{gather}\label{eq:objective_dataset}
\Lapl_{0} = \sum_{z_i \in \Set{D}_0} l\left(f_{\Set{G}}(z_i), y_i\right).    
\end{gather}
Assume $\Lapl_{0}$ is twice-differentiable and strictly convex\footnote{We provide a discussion of the non-convex and non-differential cases in Appendix~\ref{apd:non_convex}.}. Thus the hessian matrix $H_{\theta_{0}}$ is positive-definite and invertible. 
 \begin{gather}
     H_{\theta_{0}} =  \sum_{z_i \in \Set{D}_0}\nabla_{\theta_{0}}^{2} l\left(f_{\Set{G}}(z_i), y_i\right) ,
 \end{gather}
then, the final loss can be written as  
 \begin{gather}
    \hat{\theta} = \mathop{\arg\min}_{\theta}
    \Lapl, \quad
    \Lapl = \sum_{z_i \in \Set{D}_0\setminus \Delta \Set{D}} l\left(f_{\Set{G}\setminus \Delta \Set{G}}(z_i), y_i\right).
 \end{gather}
For a small perturbation $\epsilon \Lapl_{\Delta \Set{G}} $( $\epsilon$ is scalar), the parameter change $\theta_{\epsilon}$ can be expressed as
\begin{align}\label{eq:objective_TIF}
    \theta_{\epsilon} = \mathop{\arg\min}_{\theta}
    (\Lapl_0 + \epsilon \Lapl_{\Delta \Set{G}}).
\end{align}
Note that $\theta_{0}$ and $\theta_{\epsilon}$ are the minimums of equations~\eqref{eq:objective_dataset} \syd{and} \eqref{eq:objective_TIF} respectively, we have the first-order optimality conditions:
\begin{gather}
     0 = \nabla_{\theta_{\epsilon}}\Lapl_0 + \epsilon \nabla_{\theta_{\epsilon}} \Lapl_{\Delta \Set{G}}, \quad 0 =  \nabla_{\theta_0}\Lapl_0.
\end{gather}
Given that $\lim_{\epsilon \to 0} \theta_{\epsilon} = \theta_{0}$, we keep one order Taylor expansion at the point of $\theta_{0}$. Denote $\Delta \theta = \theta_{\epsilon} - \theta_{0}$, we have
\begin{align}
    0 \approx 
    \Delta \theta \left(  \nabla^{2}_{\theta_{0}}\Lapl_0 +  \epsilon \nabla^{2}_{\theta_{0}} \Lapl_{\Delta \Set{G}} \right) + \left(\epsilon \nabla_{\theta_{0}}  \Lapl_{\Delta \Set{G}} + \nabla_{\theta_0}\Lapl_0 \right).
\end{align}
Since $\Delta \Set{G}$ is a tiny subset of $\Set{G}$, we can neglect the term $\nabla^{2}_{\theta_{0}} \Lapl_{\Delta \Set{G}}$ here. With a further assumption that
\begin{align}\label{eq:tradi_esti_der}
    \text{When}~\epsilon = -1, \quad  \Lapl = \Lapl_{0} + \epsilon\Lapl_{\Delta \Set{G}},
\end{align}
% \textcolor{red}{ compared to the term $\nabla^{2}_{\theta_{0}}\Lapl_0$. Thus} 
we finally get the estimated parameter change
\begin{align}
    \hat{\theta} - \theta_{0} = \theta_{\epsilon = -1} - \theta_{0} \approx   H_{\theta_{0}}^{-1} \nabla_{\theta_{0}} \Lapl_{\Delta \Set{G}}.
\end{align}

% \frac{1}{|\Set{D}_{0}|} (\nabla_{\theta^{2}_{\theta_0}}\Lapl_0)^{-1} \nabla_{\theta_{0}} \Lapl_{\Delta \Set{G}} =

% It is worth noting that influence functions implies the two following assumptions, which is not generally established in GNN models.
% \begin{itemize}

% \item{\verb|Empirical risk minimizer|}:In reality,the model can only converge to a local minimum in most cases and therefore it is unreliable to assume that the whole minimum is reached. 

% \item{\verb|Data independence|}: The formula \par  $\mathcal{L}^\prime_{0}(Z^\prime,\theta) = \frac{1}{k-1} \sum_{i=1,i\ne j}^{k} \mathcal{L}_{0}(z_{i},\theta)$ actually implies that the training data is independent of each other,  which means that deleting or changing one or some training data will not affect the characteristics of other data samples. For example, suppose that we want to remove the influence of one node $n_{j}$ on graph $\mathcal{G}$ for a k-layer GCN model, the representation of every node $n_{j}$'s k-hop neighbors will be updated due to message passing, which represents that  $\mathcal{L}(z_{neighbor},\theta) \ne \mathcal{L}_{0}(z_{neighbor},\theta)$. Directly applying the influence functions on graph will retain the deleted node's information on graph, which means that the unlearning process is uncertified, unless removing all $n_{j}$'s adjacent nodes.

% \item{\verb|Limited in (binary) logistic regression |}
% These methods require the loss function as binary regression. And the alternative approach "one-versus-all other-classes"
% for multiple regression does not match the actual application

% \end{itemize}

% !TeX root = ./0_main.tex

\section{GIF for Graph Unlearning}
% In this section, we tailor the influence function for graph neural networks. We first identify the 
\subsection{Graph Influence Functions}
The situation where Equation~\eqref{eq:tradi_esti_der} holds is that for any sample $z_{i} \in \Set{D} \setminus \Delta\Set{D}$, its prediction from $f_{\Set{G}}$ and $f_{\Set{G} \setminus \Delta\Set{G}}$ are identical. We can formulate such an assumption as follows:
% The existence of equation~\eqref{eq:tradi_esti_der} equals 
\begin{align}\label{tradi_esi}
    \sum_{z_i \in \Set{D}_0\setminus \Delta \Set{D}} l\left(f_{\Set{G}\setminus \Delta \Set{G}}(z_i), y_i\right) = 
    \sum_{z_i \in \Set{D}_0 } l\left(f_{\Set{G}}(z_i), y_i\right) - 
    \sum_{z_i \in \Delta \Set{D}_0 } l\left(f_{\Set{G}}(z_i), y_i\right).
\end{align}
% \syd{Equation}~\eqref{tradi_esi} represents that for the loss of arbitrary $z_{i} \in \Set{D} \setminus \Delta\Set{D}$ is independent  of the diagram selection $\Set{G}$ or $\Set{G} \setminus \Delta\Set{G}$. 
This suggests that the perturbation in one sample will not affect the state of other samples.
Although it could be reasonable in the domain of text or image data, while in graph data, nodes are connected via edges by nature and rely on each other.
The neighborhood aggregation mechanism of GNNs further enhances the similarity between linked nodes. Therefore, the removal of $\Delta \Set{G}$ will inevitably influence the state of its multi-hop neighbors.
% the neighbors of nodes in $\Delta \Set{G}$ will be affected in $\Set{G}$ by message passing and are unaffected in $\Set{G} \setminus \Delta \Set{G}$ due to the removal of $\Delta \Set{G}$.
With these in mind, we next derive the graph-oriented influence function for graph unlearning in detail.
% we propose a more general and powerful graph unlearning algorithm, termed Graph-oriented Influence Function for Graph Unlearning with GIF in short. 
Following the idea of data perturbation in traditional influence function~\cite{Koh2017Understanding}, we correct Equation~\eqref{eq:objective_TIF} by taking graph dependencies into  consideration.
\begin{align}\label{eq:objective_GIF}
    \hat{\theta}_{\epsilon} = \mathop{\arg\min}_{\theta}
    (\Lapl_0 + \epsilon \Delta\Lapl_{(\Set{G}\setminus \Delta \Set{G})}), \quad
    \Delta\Lapl_{(\Set{G}\setminus \Delta \Set{G})} = \sum_{z_i \in  \Set{D}}
    \hat{l}\left(z_{i},y_{i}\right),
    \\
    \hat{l}\left(z_{i},y_{i}\right) = \begin{cases}
    l(f_{\Set{G}}(z_{i}),y_{i}),\quad & z_{i} \in \Delta\Set{G} \\
    l(f_{\Set{G}}(z_{i}),y_{i}) - 
    l(f_{\Set{G} \setminus \Delta \Set{G}}(z_{i}),y_{i}) , \quad & z_{i} \text{ is influenced by } \Delta \Set{G} \\
    0,\quad & \text{other nodes}
\end{cases} 
\end{align}

Utilizing the minimum property of $\hat{\theta}_{\epsilon}$ and denoting $\Delta \hat{\theta}_{\epsilon} = \hat{\theta}_{\epsilon} - \theta_{0}$, we similarly make a one-order expansion and deduce the answer.
\begin{align}
    0 \approx \Delta \hat{\theta}_{\epsilon} (\nabla^{2}_{\theta_{0}} \Lapl_{0} + \epsilon \nabla^{2}_{\theta_{0}} \Delta\Lapl_{(\Set{G}\setminus \Delta \Set{G})}) + (\epsilon \nabla_{\theta_{0}} \Delta\Lapl_{(\Set{G}\setminus \Delta \Set{G})} + \nabla_{\theta_{0}} \Lapl_{0}).
\end{align}
We thus have the following theorem:

\begin{theorem}
    \label{thm:GIF}(Graph-oriented Influence Functions) A general GIF algorithm for unlearning tasks has a closed-form expression:
\begin{align}\label{eq:GIF_theorem}
    \hat{\theta} - {\theta}_{0} \approx  H_{\theta_{0}}^{-1} \nabla_{\theta_{0}} \Delta\Lapl_{(\Set{G}\setminus \Delta \Set{G})}.
\end{align}
\end{theorem}

A remaining question is how large the influenced region is. To answer this question, we first define the $k$-hop neighbors of node $z_{i}$ and edge $e_{i}$ (denoted as $\Set{N}_{k}(z_{i})$ and $\Set{N}_{k}(e_{i})$, respectively) following the idea of shortest path distance (SPD). Specifically,

\begin{gather}
    \Set{N}_{k}(z_{i}) = \{ z_{j} | 1 \leqslant \text{SPD}(z_{j},z_{i}) \leqslant k \}, \\
    \Set{N}_{k}(e_{i}) = \Set{N}_{k}(z^{\prime}_{1}) \bigcup \Set{N}_{k}(z^{\prime}_{2}) \bigcup \{z^{\prime}_{1}, z^{\prime}_{2}\},
\end{gather}

\noindent where $z^{\prime}_{1}$ and $z^{\prime}_{2}$ are the two endpoints of edge $e_{i}$. Then we have the following Lemma.

\begin{lemma}
Let $A$ and $D$ denote the adjacency matrix and the corresponding degree matrix of graph $\Set{G}$. Suppose the normalized propagation matrix $\hat{A}$ is defined as $\hat{A} = D^{-\frac{1}{2}} A D^{-\frac{1}{2}}$. Then the influenced region \wrt unlearning request $\Delta \Set{G}$ for different types of graph unlearning task is:
\begin{itemize}[leftmargin=*]
    \item For Node Unlearning request $\Delta \Set{G} = \{ \Set{V}^{(rm)},\varnothing,\varnothing\}$, the influenced region is $\Set{N}_{k}(\Set{V}^{(rm)}) = \bigcup\limits_{e_{i} \in \Set{V}^{(rm)}} \Set{N}_{k+1}(e_{i})$;
    
    \item For Edge Unlearning request $\Delta \Set{G} = \{ \varnothing, \Set{E}^{(rm)}, \varnothing  \}$, the influenced region is $\Set{N}_{k}(\Set{E}^{(rm)}) = \bigcup\limits_{z_{i} \in \Set{E}^{(rm)}} \Set{N}_{k}(z_{i})$;
    
    \item For Feature Unlearning request $\Delta \Set{G} = \{ \varnothing, \varnothing, \Set{X}^{(rm)} \}$, the influenced region is $\Set{N}_{k}(\Set{X}^{(rm)}) = \bigcup\limits_{z_{i} \sim \Set{X}^{(rm)}} \Set{N}_{k}(z_{i})$, where $z_{i} \sim \Set{X}^{(rm)}$ indicates that the feature of node $z_{i}$ is revoked.
\end{itemize}

\end{lemma}

Combining the above Lemma, the formula can be further simplified to facilitate the estimation.

\begin{corollary}

Denote the prediction of $z_i$ in the remaining graph $\Set{G}\setminus \Delta \Set{G}$ using $f_{\Set{G}}$ as $f_{\Set{G}}(z_{i};\Set{G}\setminus \Delta \Set{G})$ and the parameter change as $\Delta \theta = H^{-1}_{\theta_{0}} \nabla_{\theta_{0}} \Delta\Lapl_{(\Set{G}\setminus \Delta \Set{G})}$ when $\epsilon=-1$. Then the estimated parameter change by GIF for different unlearning tasks are as follows:
\begin{itemize}[leftmargin=*]
    \item For Node Unlearning tasks,
    \begin{align}\label{eq:GIF_node}
        \Delta \theta = 
        &H_{\theta_{0}}^{-1} \sum\limits_{z_{i} \in \Set{N}_{k}(\Set{V}^{(rm)})\cup\Set{V}^{(rm)}}  \nabla_{\theta_{0}}l\left(f_{\Set{G}}(z_{i}),y_{i})\right) \nonumber\\ 
        &- H_{\theta_{0}}^{-1} \sum\limits_{z_{i} \in \Set{N}_{k}(\Set{V}^{(rm)})} \nabla_{\theta_{0}}
        l\left(f_{\Set{G}}(z_{i};\Set{G}\setminus \Delta \Set{G}),y_{i})\right).
    \end{align}
    
    \item For Edge Unlearning tasks,
    \begin{align}\label{eq:GIF_edge}
        \Delta \theta 
        =& H_{\theta_{0}}^{-1} \sum\limits_{z_{i} \in \Set{N}_{k}(\Set{E}^{(rm)})} \nabla_{\theta_{0}}
        l\left(f_{\Set{G}}(z_{i}),y_{i})\right) \nonumber\\ 
        &- H_{\theta_{0}}^{-1} \sum\limits_{z_{i} \in \Set{N}_{k}(\Set{E}^{(rm)})} \nabla_{\theta_{0}} l\left(f_{\Set{G}}(z_{i};\Set{G}\setminus \Delta \Set{G}),y_{i})\right).
    \end{align}
    
    \item For Feature Unlearning tasks,
    \begin{align}\label{eq:GIF_feature}
        \Delta \theta 
        =&
        H_{\theta_{0}}^{-1} \sum\limits_{z_{i} \sim N_{k}(\Set{X}^{(rm)})\cup\Set{X}^{(rm)}} \nabla_{\theta_{0}}
        l\left(f_{\Set{G}}(z_{i}),y_{i})\right) \nonumber\\ 
        &- H_{\theta_{0}}^{-1} \sum\limits_{z_{i} \sim N_{k}(\Set{X}^{(rm)})\cup\Set{X}^{(rm)}} \nabla_{\theta_{0}} l\left(f_{\Set{G}}(z_{i};\Set{G}\setminus \Delta \Set{G}),y_{i})\right).
    \end{align}
\end{itemize}

\end{corollary}

\subsection{Efficient Estimation}

Directly calculating the inverse matrix of Hessian matrix then computing Equation~\eqref{eq:GIF_theorem} requires the complexity of $\mathcal{O}(|\theta|^{3}+ n|\theta|^{2})$ and the memory of $\mathcal{O}(|\theta|^{2})$, which is prohibitively expensive in practice. Following the stochastic estimation method \cite{influencesta}, we can reduce the complexity and memory to $\mathcal{O}(n|\theta|)$ and $\mathcal{O}(|\theta|)$, respectively.

\begin{theorem}\label{thm:iterative_training}
    Suppose Hessian matrix $H$ is positive definite. When the spectral radius of $(I - H)$ is less than 1, $H^{-1}_{0} = I$ and $H^{-1}_{n} = ( I - H ) H^{-1}_{n-1} + I$ for $n = 1,2,3,\cdots$, then $\lim\limits_{n \to \infty} H^{-1}_{n} = H^{-1}$.
\end{theorem}

Since $\nabla_{\theta_{0}}{\Lapl_{0}}$ is frequently computed when estimating the parameter change in Equation~\eqref{eq:GIF_node} - \eqref{eq:GIF_feature}, which is however unchanged throughout the whole iteration process, we can storage it at the end of the model training once for all.
Denote $H^{-1}_{t} = \sum_{i = 0}^{t} (I - H)^{i}$, which represents the first $t$ terms in the Taylor expansion of $H^{-1}$.
Thus we can obtain the recursive equation $H^{-1}_{t} = I + (I - H)H^{-1}_{t-1}$.
Let $v = \nabla_{\theta_{0}} \Delta \Lapl_{(\Set{G} \setminus \Delta \Set{G})}$, $H_{t}^{-1}v$ be the estimation of $H_{\theta_{0}}^{-1} \nabla_{\theta_{0}}\Delta \Lapl_{(\Set{G} \setminus \Delta \Set{G})}$.
By iterating the equation until convergence, we have,
\begin{equation}\label{eq:"recursive hessian"}
    [H^{-1}_{t}v] = v + [H^{-1}_{t-1}v] - H_{\theta_{0}} [H^{-1}_{t-1}v], \quad [H^{-1}_{0}v] = v.
\end{equation}
According to Theorem \ref{thm:iterative_training}, Equation~\eqref{eq:"recursive hessian"} will naturally iterate to converge  when the spectral radius of $(I - H)$ is less than one. We take $ H^{-1}_{t}\nabla_{\theta_{0}}\Delta \Lapl_{(\Set{G} \setminus \Delta \Set{G})}$ as the estimation of $ H_{\theta_{0}}^{-1} \nabla_{\theta_{0}}{\mathcal{L}}(Z,\theta)$ . 

% \textcolor{red}{For more general hessian matrixs}, we add an extra scaling coefficient $\lambda$ for convergence and let $\lambda H_{t}^{-1}v$ be the estimation of $ H_{\theta_{0}}^{-1} \nabla_{\theta_{0}}\Delta \Lapl_{(\Set{G} \setminus \Delta \Set{G})}$ based on the recursive results.

For more general Hessian matrixes, we add an extra scaling coefficient $\lambda$ to guarantee the convergence condition. Specifically, let $\lambda H_{t}^{-1} \nabla_{\theta_{0}}\Delta \Lapl_{(\Set{G} \setminus \Delta \Set{G})}$ be the estimation based on the recursive results, then Equation~\eqref{eq:"recursive hessian"} can be modified into:
\begin{equation}\label{eq:recrusivelamda}
    [H^{-1}_{j}v] = v + [H^{-1}_{j-1}v] - \lambda H_{\theta_{0}} [H^{-1}_{j-1}v], \quad [H^{-1}_{0}v] = v.
\end{equation}
We also discuss the selection of hyperparameters $\lambda$ in Appendix~\ref{apd:ablation}.
\begin{proposition} 
The time complexity of each iteration defined in Equation~\eqref{eq:recrusivelamda} is $\mathcal{O}(|\theta|)$.
\end{proposition}

Relying on the fast Hessian-vector products (HVPs) approach \cite{fasthessian},  $H^{-1}_{\theta_{0}} [H^{-1}_{j-1}v]$  can be exactly computed in only $\mathcal{O}(|\theta|)$ time and $\mathcal{O}(|\theta|)$ memory.
% Specifically, the complexity of each iteration is $\mathcal{O}(|\theta|)$.
All computations can be easily implemented by Autograd Engine~\footnote{\url{https://pytorch.org/tutorials/beginner/blitz/autograd_tutorial.html}}. According to our practical attempts, the number of iterations $t \ll n$ in practice. The computational complexity can be reduced to $\mathcal{O}(n|\theta|)$ with $\mathcal{O}(|\theta|)$ in memory.
% 
% 
% \begin{algorithmic}
% \caption{Graph-oriented Influence functions(updating...)}
% \label{algorithm}
% \KwIn{graph $\Set{G} = \{\Set{V}, \Set{E}, \Set{X}\}$, unlearning task $\Delta \Set{G}$, scale iteration number = t, scale = $\lambda$, opitimal miminizer $\theta_{0}$} \KwOut{$\Delta \theta $}
%     \State{ nodes finding\par}
%     \State{ Forward propagation and stored the grad \par}
    
%     \State{ \text{Initialization} V = grad1 -grad2, h1 = grad1 - grad2\par}
% \For{i=1 \textbf{to} t}{
%     \State{h2 = hvps(grad0, h1)\par}
%     \State{h1 = v + h1 - h2/$\lambda$\par}
% }\EndFor
% \State{ return $\Delta \theta = h1/scale $\par}
% \end{algorithmic}
% 
% 
% 
% The complexity of each iteration is $\mathcal{O}(|\theta|)$.
% All computations can be easily implemented by Autograd Engine~\footnote{\url{https://pytorch.org/tutorials/beginner/blitz/autograd_tutorial.html}}. According to our practical attempts, the number of iterations $t \ll n$ in practice. The computational complexity can be reduced to $\mathcal{O}(n|\theta|)$ with $\mathcal{O}(|\theta|)$ in memory.

% \subsection{Application in different tasks}
% It is worth noting GIF can be applied in various tasks and models including GNN-PD models, unsupervised problem.
% In the ddl of ddl, then updating(see see if it is necessary)...

\subsection{Understanding the Mechanism of Unlearning}

To better understand the underlying mechanism of graph unlearning, we deduce the closed-form solution for one-layer graph convolution networks under a node unlearning request.
Denote $P$ and $P^{\prime}$ as the GCN model's prediction trained on the original graph and the remaining graph. And the optimal model parameters are obtained by
minimizing the following objective:
\begin{equation}
    P = \sigma(D^{-\frac{1}{2}}AD^{-\frac{1}{2}}XW), \quad {W}_{0} = \mathop{\arg\min}_{W} \Lapl_{0}.
\end{equation}

We choose Softmax as the activation function $\sigma$ and Cross entropy as measurement.
For the original graph, we denote $H = D^{-\frac{1}{2}}AD^{-\frac{1}{2}}X $. $h_{z_{i}}$ is the corresponding column of $H$ for training data $z_{i}$.
% Similarly, we have ${H}^{\prime} = {D^{\prime}}^{-\frac{1}{2}}A^{\prime}{D^{\prime}}^{-\frac{1}{2}} $ and $h^{\prime}_{z_{i}}$.
$p_{z_{i},j}$ is the predicted probability of node $z_{i}$ on the $j$-th label.
$\hat{p}_{z_{i},j} = 1 - p_{z_{i},j}$ is the error probability when the $j$-th label is the accurate label for node $z_{i}$ otherwise equals $p_{z_{i},j}$.
Similarly, we can define ${H}^{\prime}$, $h^{\prime}_{z_{i}}$, $p^{\prime}_{z_{i},j}$, and $\hat{p}^{\prime}_{z_{i},j}$ for the remaining graph.
% ${H}^{\prime} = {D^{\prime}}^{-\frac{1}{2}}A^{\prime}{D^{\prime}}^{-\frac{1}{2}} $ and $h^{\prime}_{z_{i}}$.
% We also similarly define $\hat{p}^{\prime}_{z_{i},j}$.
% $c_{i}$ is the corresponding label of node $z_{i}$.

We hope to give a closed-form solution of $\hat{W} - W_{0}$,
\begin{gather}
    P^{\prime} = \sigma({D^{\prime}}^{-\frac{1}{2}}A^{\prime}{D^{\prime}}^{-\frac{1}{2}}X^{\prime}W), \quad \hat{W} = \mathop{\arg\min}_{W} \Lapl.
    % P^{\prime} = \sigma(D^{\prime}^{-\frac{1}{2}}A^{\prime}D^{\prime}^{-\frac{1}{2}}X^{\prime}W), \quad \hat{W} = \mathop{\arg\min}_{w} \Lapl.
\end{gather}

\begin{theorem}
For one-layer GCN, the closed-form solution of $\hat{W} - {W}_{0} = ( w_{1}, w_{2},.., w_{c}), \quad w_{i} = D_{i}^{-1}(E^{(rm)}_{i} + E^{(nei)}_{i}) \in R^{F \times 1} $, where
\begin{gather}
    D_{j} = \sum_{z_{i} \in \Set{G}} p_{z_{i},j}(1-p_{z_{i},j})h_{z_i}^Th_{z_i},   \nonumber \\
    E_{j}^{(rm)} = \sum\limits_{z_{i} \in \Set{V}^{rm}} \hat{p}_{z_{i},j}h_{z_{i}}^{T}, \nonumber \quad
    E_{j}^{(nei)} = \sum\limits_{z_{i} \in \Set{N}_{K}(\Set{V}^{(rm)})} (\hat{p}_{z_{i},j}h_{z_{i}}^{T} - \hat{p}^{\prime}_{z_{i},j}{{h}^{\prime}}_{z_{i}}^{T} ).
\end{gather}
\end{theorem}

Detailed derivations are included in Appendix~\ref{sec:derivation_SGC}.

Generally, there are three key factors influencing the accuracy of parameter change estimation:

(1) $D_{j}$ represents the resistance of overall training point.
% By analyzing every part of the formula, graph-oriented influence functions shed new light on the black-box process of graph unlearning operations. 

(2) $E_{j}^{(rm)}$ represents the the influence caused by nodes from  $\Delta \Set{G}$. 

(3) $E_{j}^{(nei)}$ indicates the reflects the influence of affected neighbor nodes on parameter changes, which is jointly determined by model's accuracy and structural information in the original graph and remaining graph.

\section{Experiments}

% \subsection{Problem setup}
% For ease of display,we choose tasks on on semi-supervised node classification as an example. Major current unlearning tasks on semi-supervised node classification are proposed as followings.

% \begin{itemize}
% \item {\verb|Label Unlearning|}: simply remove some nodes out of the training set and keep these nodes on the graph
% \item{\verb|Feature/Node/Edge Unlearning|}: change some nodes' original features, delete one or some nodes or edges from graph(e.g., removal of information at the request of users)

% \end{itemize}
To justify the superiority of GIF for diverse graph unlearning tasks, we conduct extensive experiments and answer the following research questions:
\begin{itemize}[leftmargin=*]
    \item \textbf{RQ1}: How does GIF perform in terms of model utility and running time as compared with the existing unlearning approaches?
    
    \item \textbf{RQ2}: Can GIF achieve removal guarantees of the unlearned data on the trained GNN model?
    
    \item \textbf{RQ3}: How does GIF perform under different unlearning settings?
\end{itemize}

\subsection{Experimental Setup}

\begin{table}[t]
\centering
\caption{Statistics of the datasets.}
\vspace{-10pt}
\label{tab:dataset_stats}
\resizebox{0.45\textwidth}{!}{
\begin{tabular}{c|ccccc}
\hline
Dataset  & Type     & \#Nodes & \#Edges & \#Features & \#Classes \\ \hline \hline
Cora     & Citation & 2,708   & 5,429   & 1,433      & 7         \\
Citeseer & Citation & 3,327   & 4,732   & 3,703      & 6         \\
CS  & Coauthor & 18,333  & 163,788 & 6,805      & 15         \\ \hline
\end{tabular}
}
\vspace{-10pt}
\end{table}

% \usepackage{multirow}
\begin{table*}[t]
\centering
\caption{Comparison of F1 scores and running time (RT) for different graph unlearning methods for edge unlearning with 5\% edges deleted from the original graph. `LPA' and `Kmeans' stand for GraphEraser~\cite{Chen2022Graph} using balanced label propagation and balanced embedding $k$-means algorithm for community detection, respectively. The bold indicates the best result for each GNN model on each dataset.}
\label{tab:exp_overall}
\vspace{-10pt}
\begin{tabular}{cccccccc}
\hline
\multicolumn{2}{c}{Model}                             & \multicolumn{6}{c}{Dataset}                                                                                              \\ \hline
\multirow{2}{*}{Backbone} & \multirow{2}{*}{Strategy} & \multicolumn{2}{c}{Cora}               & \multicolumn{2}{c}{Citeseer}           & \multicolumn{2}{c}{CS}                 \\
                          &                           & F1 score                   & RT (second)        & F1 score                    & RT (second)        & F1 score                    & RT (second)        \\ \hline
\multirow{4}{*}{GCN}      & Retrain                   & 0.8210±0.0055          & 6.33          & 0.7318±0.0096          & 7.52          & 0.9126±0.0055          & 55.95         \\
                          & LPA                       & 0.6790±0.0001          & 1.35          & 0.5556±0.0001          & 1.69          & 0.7732±0.0001          & 3.04          \\
                          & Kmeans                    & 0.5535±0.0001          & 1.89          & 0.5045±0.0001          & 1.56          & 0.7754±0.0001          & 9.97          \\ \cline{3-8} 
                          & GIF                       & \textbf{0.8218±0.0066} & \textbf{0.16} & \textbf{0.6925±0.0060} & \textbf{0.13} & \textbf{0.9137±0.0016} & \textbf{0.26} \\ \hline
\multirow{4}{*}{GAT}      & Retrain                   & 0.8804±0.0060          & 15.72         & 0.7643±0.0049          & 19.50         & 0.9305±0.0011          & 110.39        \\
                          & LPA                       & 0.3432±0.0001          & 3.04          & 0.6997±0.0001          & 3.92          & 0.7650±0.0001          & 6.43          \\
                          & Kmeans                    & 0.6900±0.0001          & 3.19          & 0.7628±0.0001          & 3.41          & 0.8794±0.0001          & 17.36         \\ \cline{3-8} 
                          & GIF                       & \textbf{0.8649±0.0072} & \textbf{0.86} & \textbf{0.7663±0.0072} & \textbf{0.59} & \textbf{0.9325±0.0015} & \textbf{1.02} \\ \hline
\multirow{4}{*}{SGC}      & Retrain                   & 0.8236±0.0142          & 6.63          & 0.7132±0.0091          & 7.16          & 0.9165±0.0045          & 58.12         \\
                          & LPA                       & 0.3247±0.0001          & 2.03          & 0.3934±0.0001          & 1.70          & 0.5267±0.0001          & 3.08          \\
                          & Kmeans                    & 0.3690±0.0001          & 1.41          & 0.3874±0.0001          & 1.56          & 0.6532±0.0001          & 10.59         \\ \cline{3-8} 
                          & GIF                       & \textbf{0.8129±0.0110} & \textbf{0.12} & \textbf{0.6892±0.0082} & \textbf{0.12} & \textbf{0.9164±0.0051} & \textbf{0.24} \\ \hline
\multirow{4}{*}{GIN}      & Retrain                   & 0.8051±0.0144          & 8.48          & 0.7294±0.0207          & 9.94          & 0.8822±0.0074          & 70.38         \\
                          & LPA                       & 0.6605±0.0001          & 2.65          & 0.6096±0.0001          & 2.21          & 0.6510±0.0001          & 3.82          \\
                          & Kmeans                    & 0.7491±0.0001          & 2.53          & 0.6517±0.0001          & 2.11          & 0.8336±0.0001          & 10.29         \\ \cline{3-8} 
                          & GIF                       & \textbf{0.8059±0.0234} & \textbf{0.48} & \textbf{0.7315±0.0185} & \textbf{0.48} & \textbf{0.8884±0.0083} & \textbf{0.57} \\ \hline
\end{tabular}
\end{table*}

\vspace{5pt}\noindent\textbf{Datasets}. We conduct experiments on three public graph datasets with different sizes, including Cora~\cite{Kipf2017GCN}, Citeseer~\cite{Kipf2017GCN}, and CS~\cite{Chen2022Graph}. These datasets are the benchmark dataset for evaluating the performance of GNN models for node classification task. Cora and Citeseer are citation datasets, where nodes represent the publications and edges indicate citation relationship between two publications, while the node features are bag-of-words representations that indicate the presence of keywords. CS is a coauthor dataset, where nodes are authors who are connected by an edge if they collaborate on a paper, the features represent keywords of the paper. For each dataset, we randomly split it into two subgraphs --- a training subgraph that consists of 90\% nodes for model training and a test subgraph containing the rest nodes for evaluation. The
statistics of all three datasets are summarized in Table~\ref{tab:dataset_stats}.

\vspace{5pt}\noindent\textbf{GNN Models}. We compare GIF with four widely-used GNN models, including GCN~\cite{Kipf2017GCN}, GAT~\cite{GAT}, GIN~\cite{GIN}, and SGC~\cite{SGC}.

% For each GNN model, we stack two layers of GNN models.
% Since the approximation of traditional influence functions did not consider the dependencies between nodes, which the attack can identify the presence of unlearning data due to the information residual of the local neighborhood. We remove both the deleted and affected nodes from graph. Besides, since traditional influence functions require the loss function as (binary) logistic regression, we use the one-vs-rest strategy splits the multi-class classification into one binary classification
% problem per class and train with (binary) logistic regression."

\vspace{5pt}\noindent\textbf{Evaluation Metrics}. We evaluate the performance of GIF in terms of the following three criteria:

\begin{itemize}[leftmargin=*]

\item {\verb|Unlearning Efficiency|}: 
We record the running time to reflect the unlearning efficiency across different unlearning algorithms.

\item{\verb|Model Utility|}: 
As in GraphEraser~\cite{Chen2022Graph}, we use F1 score --- the harmonic average of precision and recall --- to measure the utility.

\item{\verb|Unlearning Efficacy|}: 
Due to the nonconvex nature of graph unlearning, it is intractable to measure the unlearning efficacy through the lens of model parameters. Instead, we propose an indirect evaluation strategy that measures model utility in the task of forgetting adversarial data. See section~\ref{ssec:exp_attack} for more details.

% \item{\verb|Universality|}:
% Some unlearning methods  are limited to linear GNN\cite{grapheditor,graphcerti} or limited tasks, which represent poor universality. Therefore, We propose this metric to measure even if such algorithms can be widely used in practice by mainstream models.

\end{itemize}

\vspace{5pt}\noindent\textbf{Baselines}.
We compare the proposed GIF with the following unlearning approaches:
\begin{itemize}[leftmargin=*]
    \item{\verb|Retrain|}. This is the most straightforward solution that retrains GNN model from scratch using only the remaining data. It can achieve good model utility but falls short in unlearning efficiency.
    
    \item{\verb|GraphEraser|}~\cite{Chen2022Graph}. This is an efficient retraining method. It first divides the whole training data into multiple disjoint shards and then trains one sub-model for each shard. When the unlearning request arrives, it only needs to retrain the sub-model of shards that comprise the unlearned data. The final prediction is obtained by aggregating the predictions from all the sub-models. GraphEraser has two partition strategies for shard splitting, namely balanced LPA and balanced embedding $k$-means. We consider both of them as baselines.
    \item{\verb|Influence Function|}~\cite{Koh2017Understanding}. It can be regarded as a simplified version of our GIF, which only considers the individual sample, ignoring the interaction within samples. 
\end{itemize}

It's worth mentioning that we do not include some recent methods like GraphEditor~\cite{Cong2022GRAPHEDITOR} (not published yet) and \cite{certified} since they are tailored for linear models. 

\noindent\textbf{Implementation Details}.
We conduct experiments on a single GPU server with Intel Xeon CPU and Nvidia RTX 3090 GPUs.
In this paper, we focus on three types of graph unlearning tasks.
For node unlearning tasks, we randomly delete the nodes in the training graph with an unlearning ratio $\rho$, together with the connected edges.
For edge unlearning tasks, each edge on the training graph can be revoked with a probability $\rho$.
For feature unlearning tasks, the feature of each node is dropped with a probability $\rho$, while keeping the topology structure unchanged.
All the GNN models are implemented with the PyTorch Geometric library. Following the settings of GraphEraser~\cite{Chen2022Graph}, for each GNN model, we fix the number of layers to 2, and train each model for 100 epochs. For GraphEraser, as suggested by the paper~\cite{Chen2022Graph}, we partition the three datasets, Cora, Citeseer, and Physics, with two proposed methods, BLPA and BKEM, into 20, 20, and 50 shards respectively. For the unique hyperparameters of GIF, we set the number of iterations to 100, while tuning the scaling coefficient $\lambda$ in the range of $\{ 10^{1}, 10^{2}, 10^{3}, 10^{4}, \cdots\}$.
All the experiments are run 10 times. We report the average value and standard deviation.

\subsection{Evaluation of Utility and Efficiency (RQ1)}
To answer \textbf{RQ1}, we implement different graph unlearning methods on four representative GNN models and test them on three datasets.
Table~\ref{tab:exp_overall} shows the experimental result for edge unlearning with ratio $\rho=0.05$. We have the following \textbf{Obs}ervations:

\begin{itemize}[leftmargin=*]
\item \syd{\textbf{Obs1: LPA and Kmeans have better unlearning efficiency than Retrain while failing to guarantee the model utility.}
Firstly, we can clearly observe that Retrain can achieve excellent performance in terms of the model utility.
However, it suffers from worse efficiency with a large running time.
Specifically, on four GNN models, Retrain on the CS dataset is approximately 5 to 20 times the running time of other baseline methods.
Meanwhile, other baseline methods also cannot guarantee the performance of the model utility.
For example, when applying LPA to different GNN models on Cora dataset, the performance drops by approximately 17.9\%$\sim$60.5\%, compared with Retrain.
For Kmeans, it will drop by around 15.0\%$\sim$28.72\% on the Citeseer dataset with four different GNN models.
These results further illustrate that existing efforts cannot achieve a better trade-off between the unlearning efficiency and the model utility.}
\item \syd{\textbf{Obs2: GIF consistently outperforms all baselines and can achieve a better trade-off between unlearning efficiency and model utility.} For the performance of node classification tasks, GIF can completely outperform LPA and Kmeans, and even keep a large gap.
For example, on the Cora dataset, GIF outperforms LPA by around 21.0\%$\sim$150.4\% over 4 different GNN models, and outperforms Kmeans by 7.58\%$\sim$120.3\%.
In addition, GIF greatly closes the gap with Retrain.
For instance, GIF achieves a comparable performance level with Retrain on all datasets over 4 different GNN models, and it even achieves better performance than Retrain with GCN models on Cora and CS datasets.
These results demonstrate that GIF can effectively improve the performance of the model utility.
In terms of unlearning efficiency, GIFs can effectively reduce running time.
Compared with LPA and Kmeans, GIF can shorten the running time by about 15$\sim$40 times.
Moreover, compared with Retrain, GIF can even shorten the running time by hundreds of times, such as GAT on the CS dataset.
These results fully demonstrate that GIF can guarantee both performance and efficiency, and consistently outperform existing baseline methods.
GIF effectively closes the performance gap with Retrain and achieves a better trade-off between unlearning efficiency and the model utility.}
\end{itemize}

\begin{figure*}[t]
\centering
\includegraphics[width=0.8\linewidth]{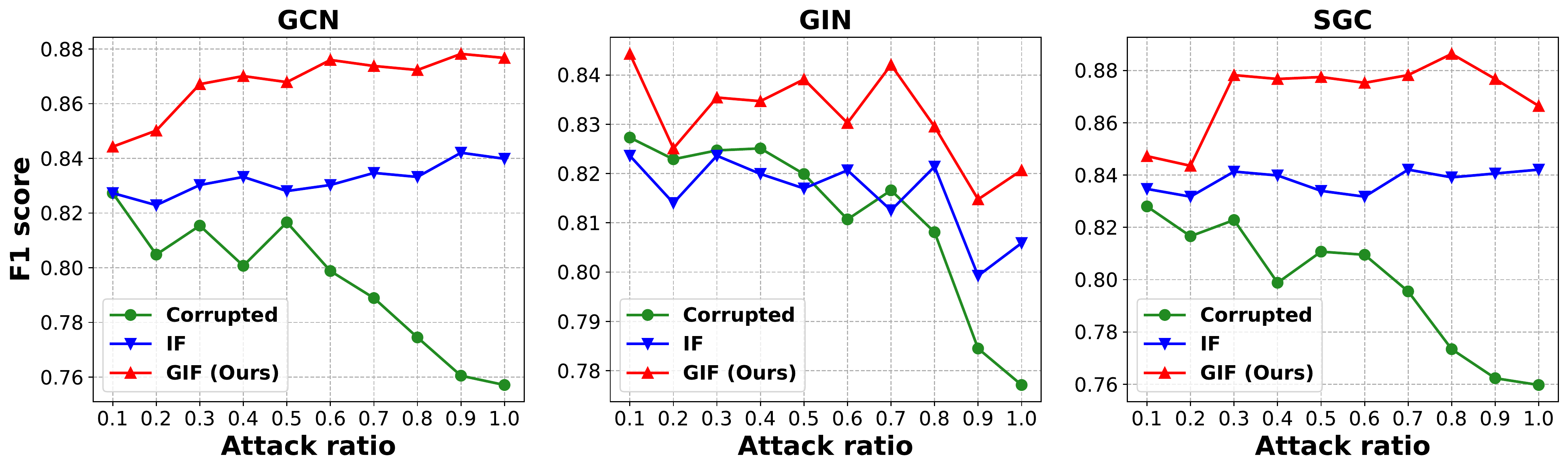}
\vspace{-10pt}
\caption{Comparison of unlearning efficacy over three GNN models.}
\Description[Performance comparison]{Comparison of unlearning efficacy over three GNN models.}
\label{fig:attack}
% \vspace{-4mm}
\end{figure*}

\subsection{Evaluation of Unlearning Efficacy  (RQ2)}\label{ssec:exp_attack}

The paramount goal of graph unlearning is to eliminate the influence of target data on the deployed model. However, it is insufficient to measure the degree of data removal simply based on the model utility. Given that IF-based methods, including GIF and traditional IF, generate the new model by estimating parameter changes in response to a small mass of perturbation, a tricky unlearning mechanism may deliberately suppress the amount of parameter change, thus achieving satisfactory model utility.
To accurately assess the completeness of data removal, we propose a novel evaluation method inspired by adversarial attacks. 
Specifically, we first add a certain number of edges to the training graph, satisfying that each newly-added edge links two nodes from different classes. The adversarial noise will mislead the representation learning, thus reducing the performance.
Thereafter, we use the adversarial edges as unlearning requests and evaluate the utility of the estimated model by GIF and IF. Intuitively, larger utility gain indicated higher unlearning efficacy. Figure~\ref{fig:attack}
shows the F1 scores of the different models produced by GIF and IF under different attack ratios on Cora. We have the following \textbf{Obs}ervations:
\begin{itemize}[leftmargin=*]
\item \textbf{Obs3: Both GIF and IF can reduce the negative impact of adversarial edges, while GIF consistently outperforms IF.}
The green curves show the performance of directly training on the corrupted graphs.
We can clearly find that as the attack ratio increases, the performance shows a significant downward trend.
For example, on GCN model, the performance drops from 0.819 to 0.760 as the attack ratio increases from 0.5 to 0.9; on SGC model, the performance drops from 0.817 to 0.785 as the attack ratio increases from 0.7 to 0.9.
These results illustrate that adversarial edges can greatly degrade performance.
The results of IF are shown as blue curves.
We can observe that the trend of performance degradation can be significantly alleviated.
For example, on GCN model, compared with training on the corrupted graphs, when the attack ratio is increased from 0.5 to 0.9, the performance is relatively increased by 5\%$\sim$10\%.
On SGC, there are also relative improvements of 5\%$\sim$10\% on performance when the attack ratio is increased from 0.7 to 0.9.
These results demonstrate that IF can effectively improve the unlearning efficacy.
Finally, we plot the experimental results of GIF with the red curves.
We observe that GIF consistently outperforms other methods and maintains a large gap.
Compared with IF, the performance is improved by about 2.1\%$\sim$4.7\% on the GCN model and is improved by about 1\%$\sim$5\% on SGC model.
In addition, except for the GIN model, the performance of GIF even improves as the ratio goes up.
These results further demonstrate that GIF can effectively improve the unlearning efficacy by considering the information of graph structure.
% \item \textbf{Obs4: GIF performs better as attack ratio increases.}
\end{itemize}

\subsection{Hyperparameter Studies (RQ3)}
\begin{figure*}[t]
\centering
\includegraphics[width=0.85\linewidth]{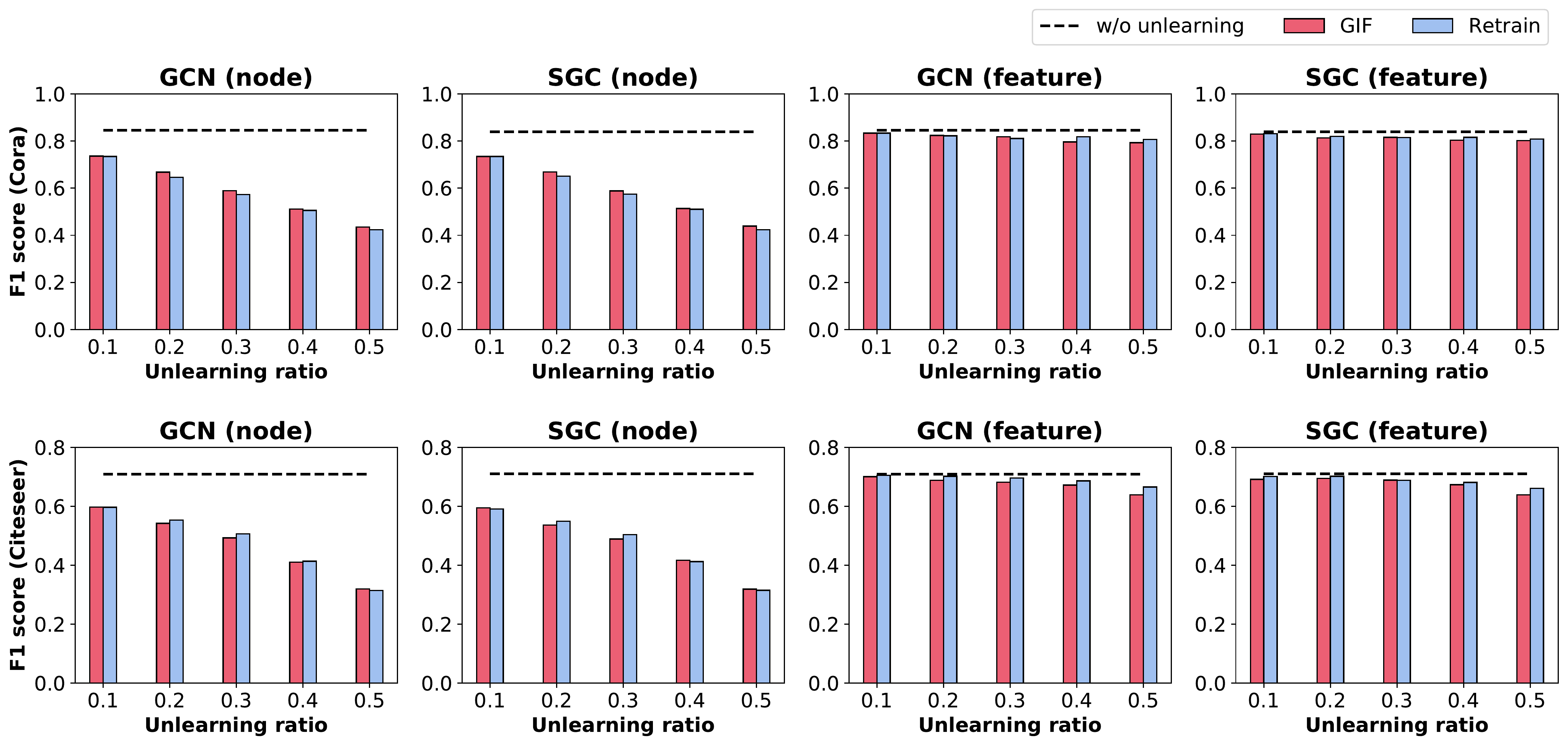}
\vspace{-10pt}
\caption{Impact of unlearning ratio $\rho$ on node unlearning tasks and feature unlearning tasks.}
\Description[Impact of unlearning ratio]{Node unlearning task and feature unlearning task.}
\label{fig:unlearning_ratio}
% \vspace{-4mm}
\end{figure*}
We then move on to studying the impact of hyperparameters.
We implement two unlearning tasks with different unlearning ratios $\rho$ on two representative GNN models, GCN and SGC, to compare the performance of the GIF with the retrained model.
We also study the impact of scaling coefficient $\lambda$ in Appendix D.
Figure~\ref{fig:unlearning_ratio} shows the experimental result of the node unlearning and feature unlearning with ratio $\rho$ range from $0.1$ to $0.5$ on Cora and Citeseer datasets. We have the following \textbf{Obs}ervations:
\begin{itemize}[leftmargin=*]
\item \textbf{Obs4: GIF consistently achieves comparable performance to Retrain on both node and feature unlearning tasks.} 
The black dashed lines represent the performance without unlearning, which is an upper bound on performance.
From the results, we can clearly observe that GIF can achieve comparable performance to Retrain on both node unlearning and feature unlearning tasks, with different models or datasets.
Specifically, for the node unlearning tasks, the performance of both methods gradually decreases as the ratio increases, in contrast to feature unlearning tasks.
Nonetheless, GIF can achieve a similar performance compared to Retrain, and even surpass Retrain on the Cora dataset.
These results illustrate that GIF can efficiently handle node-level tasks and can replace Retrain-based methods.
For feature unlearning tasks, we can observe that the performance drops less compared to node unlearning tasks.
Since node unlearning tasks need to remove the local structure of the entire graph, including node features and neighbor relationships. 
While feature unlearning tasks just need to remove some node features, which will lead to less impact on performance.
From the results, we can find that GIF can also achieve comparable performance to retraining on different datasets or GNN models.
These experimental results and analyses further demonstrate that GIF can effectively handle both node-level and feature-level tasks, and can replace retraining-based methods to achieve outstanding performance.
    % \item For the node unlearning tasks, though the F1 score descending with the unlearning score, our GIF method performs almost the same as the Retrain, which means that GIF does not impair the effectiveness of the model. 
    % \item For the feature unlearning tasks, the F1 score hardly influenced by the unlearning ratio and our GIF is comparable with the retraining model.
\end{itemize}
% !TeX root = ./0_main.tex
\section{Related Work}

\noindent\textbf{Machine unlearning} aims to eliminate the influence of a subset of the training data from the trained model out of privacy protection and model security. Ever since Cao $\&$ Yang~\cite{Cao&Yang2015} first introduced the concept, several methods are proposed to address the unlearning tasks, which can be classified into two branches: exact approaches and approximate approaches.

\noindent\textbf{Exact approaches.}
Exact unlearning methods aim to create models that perform identically to the model trained without the deleted data, or in other words, retraining from scratch, which is the most straightforward way but computationally demanding. Many prior efforts designed models, say Ginart \etal~\cite{Ginartetal2019} described unlearning approaches for k-means clustering while Karasuyama \etal~\cite{M.Karasuyamaetal2010} for support vector machines. Among these, the $SISA$(sharded, isolated, sliced, and aggregated) approach~\cite{machine} partitions the data and separately trains a set of constituent models, which are afterwards aggregated to form a whole model. During the procedure of unlearning, only the affected submodel is retrained smaller fragments of data, thus greatly enhance the unlearning efficiency. Follow-up work GraphEraser~\cite{Chen2022Graph} extends the shards-based idea to graph-structured data, which offers partition methods to preserve the structural information and also designs a weighted aggregation for inference. But inevitably, GraphEraser could impair the performance of the unlearned model to some extent, which are shown in our paper.

\noindent\textbf{Approximate approaches.}
Approximate machine unlearning methods aim to facilitate the efficiency through degrading the performance requirements, in other words, to achieve excellent trade-offs between the efficiency and effectiveness. Recently, influence function ~\cite{Koh2017Understanding} is proposed to measure the impact of a training point on the trained model. Adapting the influence function in the unlearning tasks, Guo \etal~\cite{certified} introduced a probabilistic definition of unlearning motivated by differential privacy~\cite{DPDwork11}, and proposed to unlearn by removing the influence of the deleted data on the model parameters. Specifically, they used the deleted data to update ML models by performing a Newton step to approximate the influence of the deleted data and  remove it, then they introduced a random noise to the training objective function to ensure the certifiability. With a similar idea, Izzo \etal~\cite{Izzoetal2021} propose an approximate data deletion method with a computation time independent of the size of the dataset.
Another concurrent work~\cite{chenICLRinfluence} proposes a similar formula for edge and node unlearning tasks on simple graph convolution model, and further analyzes the theoretical error bound of the estimated influences under the Lipschitz continuous condition.

%\subsection{Graph Unlearning}
%\cite{graph} proposes to split the original graph into balanced shards by BLPA or BEKM Algorithms and  aggregate all shards by a learning-based method. When receiving unlearning tasks, the model provider can simply retrain the corresponding shards and aggregate them again. \cite{grapheditor} updates parameters for SGC by solving a Ridge regression problem $L_{Ridge}(W;X,Y) = \lVert{XW - Y} \rVert_{F}^{2} + \lambda \lVert{W}\rVert_{F}^{2}$

% !TeX root = ./0_main.tex

\section{Conclusion and Future work}
% \textbf{Other GNN domains}
% It is worth our proposed GIF is not limited in the domain of graph unlearning. For example, GIF can be applied to find a subgraph structure for the graph
% prediction
% Motivated by actively, unlearn some elements on graph or subgraph structure which has the greatest impact on several data or model predictions utilized by Monte Carlo Tree Search\cite{Mentocarlosuvery}(see Appendix C for more detail). GIF can also be used for the purpose like detecting the adversarial data-poisoning attack.

In this work, we study the problem of graph unlearning that removes the influence of target data from the trained GNNs. We first unify different types of graph unlearning tasks \wrt node, edge, and feature into a general formulation. Then we recognize the limitations of existing influence function when solving graph data and explore the remedies. In particular, we design a graph-oriented influence function that considers the structural influence of deleted nodes/edges/features on their neighbors. Further deductions on the closed-form solution provide a better understanding of the unlearning mechanism.
We conduct extensive experiments on four GNN models and three benchmark datasets to justify the advantages of GIF for diverse graph unlearning tasks in terms of unlearning efficacy, model utility, and unlearning efficiency.

At present, the research on graph unlearning is still at an early stage and there are several open research questions.
In future work, we would like to explore the potential of graph influence function in other applications, such as recommender systems~\cite{abs-2210-11054,abs-2201-02327}, influential subgraph discovery, and certified data removal. Going beyond explicit unlearning requests, 
we will focus on the auto-repair ability of the deployed model --- that is, to discover the adversarial attacks to the model and revoke their negative impact. Another promising direction is the explainability of unlearning algorithms that can improve the experience in human-AI interaction.

\noindent\textbf{Acknowledgement}.
This work is supported by the National Key Research and Development Program of China (2020AAA0106000), the National Natural Science Foundation of China (9227010114, U21B2026), and the CCCD Key Lab of Ministry of Culture and Tourism.

\bibliographystyle{ACM-Reference-Format}
\balance
\bibliography{myref}

\newpage
% !TeX root = ./0_main.tex
\clearpage
\appendix

% \section{Discussion}\label{sec:discussion}

% \textbf{Convex Damage}
% It is worth noting that most current GNN models are shallow, layers.  Several studies  in other domains also show that influence  functions in non-convex shallow networks can  achieve excellent results, as  CNN\cite{Influences},\cite{fragile}. However,  when the number of layers is too deep, the prediction will become inaccurate due to the convex nature of loss function will be broken very heavily(i.e., the curvature value of the network at optimal model parameters might be quite large\cite{fragile}). Thus, algorithms related to influence functions in deep networks\cite{jknet} are still a unexplored territory.

\section{non-convex and non-differential discussion \label{apd:non_convex}}
% It is worth noting that most current GNN models are shallow, layers.
Prior studies in other domains show that influence functions in non-convex shallow networks can achieve excellent results, such as  CNN\cite{Koh2017Understanding}. However, when the number of layers goes too deep, the prediction will become inaccurate since the convexity of the objective function will be broken heavily (\ie the curvature value of the network at optimal model parameters might be quite large\cite{fragile}). \cite{Koh2017Understanding} proposes a convex quadratic approximation of the loss around the optimal parameter $\theta$ and smooth approximations to non-differentiable losses. Concurrent work\cite{chenICLRinfluence} provides some error analysis based on simple graph convolutions. Thus, strict and general error analysis related to influence functions in deep networks\cite{jknet} is still an unexplored territory.

\section{proof of Theorem \ref{thm:iterative_training}}\label{sec:proofof5}
% \begin{theorem}
% Hessian matrix $H$ is positive define(PD). When the spectral radius of $(I - H)$ is less than one, $H^{-1}_{0} = I, \quad H^{-1}_{n} = ( I - H ) H^{-1}_{n-1} + I, n = 1,2,3,..., \quad \text{then} \lim\limits_{n \to \infty} H^{-1}_{n} = H^{-1}$
% \end{theorem}
Consider the $n$-th term of $H^{-1}_{n}$,
\begin{align}
    &H^{-1}_{n} = ( I - H ) H^{-1}_{n-1} + I \\
    \Rightarrow & H^{-1}_{n} - H^{-1} = ( I - H ) (H^{-1}_{n-1} - H^{-1})\\
    \Rightarrow & H^{-1}_{n} = (I - H)^{n} + H^{-1}.
\end{align}
here $(I - H)$ is a square $|\theta| \times |\theta|$  matrix with $|\theta|$ linearly independent eigenvectors $V_{i} \in \Space{R}^{|\theta| \times 1}$ and the corresponding eigenvalue $\lambda_{i}$ (where $i = 1, ..., |\theta|$). Then it can be factorized as
\begin{align}
    (I - H ) = T U T^{-1},
\end{align}
where $T$ is a square $|\theta| \times |\theta|$ matrix whose $i$-th column is the eigenvector $v_{i}$, and $U$ is the diagonal matrix whose diagonal elements are the corresponding eigenvalues, $U_{ii} = \lambda_{i}$
\begin{align}
    (I - H )^{n} = T U^{n} T^{-1},
\end{align}
$T$ and $T^{-1}$ are fixed value matrix. Since $\forall |\lambda_{i}| < 1$, denote $|\lambda| = \max\limits_{1\leqslant i \leqslant |\theta|} |\lambda_{i}|$. 
\begin{align}
    \forall 1 \leqslant s,t \leqslant |\theta|, |u_{st}|^{n} \leqslant |\lambda|^{n}, \quad
    \lim_{n \to \infty} |\lambda|^{n} = 0,
\end{align}
\begin{align}
     \text{Thus,} \quad \lim_{n \to \infty} U^{n} = 0 
     \Rightarrow \lim_{n \to \infty}(I - H)^{n} = \lim_{n \to \infty} T U^{n} T^{-1} = 0. \\
\end{align}
And therefore, we prove that $\lim_{n \to \infty}H^{-1}_{n} = H^{-1}$.

\section{Derivation of one-layer GCN model}\label{sec:derivation_SGC}

% \begin{theorem}
% For one layer graph convolution network, the closed-form solution of $\hat{W} - {W}_{0} = ( w_{1},  w_{2},..,  w_{c}), w_{i} = D_{i}^{-1}(E^{(rm)}_{i} + E^{(nei)}_{i}) \in R^{F \times 1} $
% \begin{align}
%     D_{j} = \sum_{z_{i} \in \Set{G}} p_{z_{i},j}(1-p_{z_{i},j})h_{z_i}^Th_{z_i},   \nonumber \nonumber, \quad
%     E_{j}^{(rm)} = \sum\limits_{z_{i} \in \Set{V}^{rm}} \hat{p}_{z_{i},j}h_{z_{i}}^{T}, \nonumber\\
%     E_{j}^{(nei)} = \sum\limits_{z_{i} \in N_{K}(V^{rm})} (\hat{p}_{z_{i},j}h_{z_{i}}^{T} - \hat{p}^{\prime}_{z_{i},j}{h}^{\prime}_{z_{i}}^{T} )
% \end{align}
% \end{theorem}
We first calculate the first-order gradient of the arbitrary element in matrix $P_{log}$ in Corollary 8, then we derive its Hessian matrix in Corollary 10, finally we substitute the results into Equation (20) and finish the proof.

Notations: The number of nodes is  $M = |\Set{V}|$ with $C$ classes.
For $\odot$, $A,B \in \Space{R}^{s \times  t}, A \odot B = (a_{ij}b_{ij})_{s \times  t} \in \Space{R}^{s \times  t}$. $\otimes$ is the Kronecker product. Denote $Vec(A)$ as the vectorization of matrix $A$, which is a linear transformation which converts the matrix into a column vector.$P = (p_{mn})_{M\times C},  P_{log} = (l_{mn})_{M\times C}, T = (t_{mn})_{M \times C}, $. $P_{log}$ is equivalent to take the logarithm of each element of the matrix $P$. 
$I_{c\times c} \in \Space{R}^{C \times C} $ is an all-one matrix.  $I^{(M_{i},N_{i})}_{M\times N} = (i_{ij})_{M \times N} \in \Space{R}^{M \times N}$, $i_{M_{i},N_{i}} = 1$ and the rest are all zero. $I_{c} \in \Space{R}^{C \times C}$ is an identity matrix. $H = [h1,h2,..,h_{M}]^{T}, \forall h_{j} \in \Space{R}^{1 \times F}$,
$W = (w_{ij})_{F \times C} \in \Space{R}^{F\times C}, \quad \forall 1\leqslant M_{i} \leqslant M, 1\leqslant C_{i} \leqslant C$
We define $\nabla_{W} l_{M_{i},C_{i}} \in \Space{R}^{F \times C},$ whose $s$-th and $t$-th column element is $
    \quad (\nabla_{W} l_{M_{i},C_{i}})_{st} = \frac{\partial l_{M_{i},C_{i}}}{\partial w_{st}}.
$

We now deduce the first order gradient.
\begin{corollary}\label{cor:"one order"}
For any element $l_{M_{i},C_{i}}$  of matrix $P_{log}$,  we have 
    $\nabla_{W}l_{M_{i},C_{i}} = (H)^{T}[I_{M\times C}^{(M_i,C_i)}-I_{M\times M}^{(M_i,M_i)}Softmax(T)]$
\end{corollary}
Proof:
\begin{align}\label{eq:'matrixdone'}
    P_{log} = LogSoftmax(T), \quad
     \Rightarrow dP_{log} = dLogSoftmax(T),
\end{align}
\begin{align}
    \Rightarrow l_{mn} = log(\frac{e^{t_{mn}}}{\sum_{i=1}^C e^{t_{mi}}}) \quad \Rightarrow dl_{mn}= dt_{mn} - \frac{\sum_{i=1}^C (e^{t_{mi}}dt_{mi})}{\sum_{i=1}^C e^{t_{mi}}}.
\end{align}
Substitute the equation above  back into ~\eqref{eq:'matrixdone'}
\begin{align}\label{eq:"dequation"}
    dP_{log} = dT - (Softmax(T) \odot dT)I_{C \times C}.
\end{align}
For training point $z_{i}$, denote the prediction of the corresponding label in $P$ is $p_{M_{i},C_{i}}$. With expectation to decompose the left side to the following form
\begin{align}
    dl_{M_{i},C_{i}} = \nabla_{W}l_{M_{i},C_{i}}\odot dW.
\end{align}
We first decompose it into 
\begin{align}
    dl_{M_{i},C_{i}} = \nabla_{T}l_{M_{i},C_{i}}\odot dT.
\end{align}
Calculating the corresponding term in equation~\eqref{eq:"dequation"},
\begin{align}
    \nabla_{T}l_{M_{i},C_{i}}  = I_{M\times C}^{(M_i,C_i)}-I_{M\times M}^{(M_i,M_i)}Softmax(T)
\end{align}

\begin{align}
    \text{Since} \quad T = HW  \quad
    \Rightarrow \nabla_{W}l_{M_{i},C_{i}} = H^{T}\nabla_{T}l_{M_{i},C_{i}}
\end{align}

\begin{align}
    \Rightarrow \nabla_{W}l_{M_{i},C_{j}} = (H)^{T}[I_{M\times C}^{(M_i,C_i)}-I_{M\times M}^{(M_i,M_i)}Softmax(T)]
\end{align}

\begin{align}
    \nabla_{Vec(W)}l_{M_{i},C_{i}} = Vec((H)^{T}\{I_{M\times C}^{(M_i,C_i)}-I_{M\times M}^{(M_i,M_i)}Softmax(T)\})
\end{align}

In what follows, we derive the Hessian matrix. For ease of notation, we denote $\nabla_{Vec(W)}l_{M_{i},C_{i}} = Vec(\nabla_{W}l_{M_{i},C_{i}})$, the $i$-th element of $Vec(W)$ is $w^{(v)}_{i}$.

\begin{lemma}\label{the:"softmaxder}
    For $\forall T \in \Space{R}^{M \times C}$, we have
    \begin{align}
    dSoftmax(T) = 
    &Softmax(T)\odot dT \nonumber \\
    &-Softmax(T)\odot [(Softmax(T)\odot dT)I_{C\times C}]
    \end{align}
\end{lemma}

\begin{corollary}\label{cor:"two order"}
For $\forall 1\leqslant M_{i} \leqslant M, 1\leqslant C_{i} \leqslant C$, we define $\nabla^{2}_{W} l_{M_{i},C_{i}} \in \Space{R}^{FC \times FC}$, whose $s$-row $t$-column element is $
    (\nabla^{2}_{W} l_{M_{i},C_{i}})_{st} = \frac{\partial^{2} l_{M_{i},C_{i}}}{\partial w^{(v)}_{s} \partial w^{(v)}_{t}}.$
Then, $\nabla^2_{W}l_{M_{i},C_{i}}$ can be block into the following form:
\begin{align}
    \nabla^2_{W}l_{M_{i},C_{i}} = diag\{D_{1}^{(i)},D_{2}^{(i)},...,D_{c}^{(i)}\},\quad  D_{j}^{(i)} = p_{M_i,j}(1-p_{M_i,j})h_{M_i}^Th_{M_i}
\end{align}
    
\end{corollary}

Proof:
\begin{align}
    d\nabla_{vec(W)}l_{M_{i},C_{i}} = -dVec((\widehat{H})^T(I_{M\times M}^{(M_i,M_i)}Softmax(T))
\end{align}

\begin{align}
    &\Rightarrow d\nabla_{vec(W)}l_{M_{i},C_{i}} = -(I_C \otimes (\widehat{H})^T)Vec(d(I_{M\times M}^{(M_i,M_i)}Softmax(T))
    \\
    &\Rightarrow d\nabla_{vec(w)}l_{M_{i},C_{i}} = -(I_C \otimes (\widehat{H})^T)(I_C \otimes I_{M\times M}^{(M_i,M_i)})Vec(dSoftmax(T))
\end{align}

Utilizing Lemma~\ref{the:"softmaxder}, we have
\begin{align}
    Vec(dSoftmax(T))= 
    &diag(Softmax(T))  \nonumber \\ 
    &\cdot[I-(I_{C\times C}\odot I_M)diag(SoftmaxT)]\nonumber\\
    &\cdot(I_{C\times C}\odot H)Vec(dW)
\end{align}

\begin{gather}
    \Rightarrow d\nabla_{vec(W)}l_{M_{i},C_{i}} = -(I_C \otimes (\widehat{H})^T)(I_C \otimes I_{M\times M}^{(M_i,M_i)})\cdot\nonumber\\diag(Softmax(T)\{I-(I_{C\times C}\odot I)diag(SoftmaxA)\}(I_C\odot H)Vec(dW)
\end{gather}
\begin{align}
   \Rightarrow \nabla^2_{W}l_{M_{i},M_{i}} = -(I_C \otimes (\widehat{H})^T)(I_C \otimes I_{M\times M}^{(M_i,M_i)})\nonumber\\diag(Softmax(T)\{I_{M\times C}-diag(Softmax(T)\}(I_C\otimes H) 
\end{align}

Then we multiply the corresponding matrices together and simplify the result.
\begin{gather}
    \Rightarrow \nabla^2_{W}l_{M_{i},C_{i}} = diag\{D_{1}^{(i)},D_{2}^{(i)},...,D_{C}^{(i)}\}\cdot
    \\ D_{j}^{(i)}=H^TI_{M\times M}^{(M_i,M_i)}\tilde{P}_{j}(1-\tilde{P}_{j})H = p_{M_i,j}(1-p_{M_i,j})h_{M_i}^Th_{M_i}
\end{gather}

 We then have
\begin{align}
    H_{\theta_{0}} = diag\{D_{1},D_{2},...,D_{C}\}, \quad D_{j} = \sum_{z_{i} \in \Set{G}} p_{M_i,j}(1-p_{M_i,j})h_{M_i}^Th_{M_i}
\end{align}
\begin{align}
    \nabla_{\theta_{0}} \Delta\Lapl_{(\Set{G}\setminus \Delta \Set{G})} =
    &Vec( \sum_{z_{i} \in \Set{V}^{rm} } \nabla_{W_{0}}l_{M_{i},C_{i}} \nonumber \\ 
    &+
    \sum_{z_{i} \in {N_{K}}(\Set{V}^{rm})}
    ( \nabla_{W_{0}}l_{M_{i},C_{i}} -  \nabla_{W_{0}}\hat{l}_{M_{i},C_{i}} ) 
\end{align}
\begin{align}
     H^{T} P_{z_{i}} = (\hat{P}_{M_{i},C_{1}}h_{M_{i}}^{T}, \hat{P}_{M_{i},C_{2}}h_{M_{i}}^{T},..,\hat{P}_{M_{i},c_{c}}h_{M_{i}}^{T}) \in \Space{R}^{F \times C}
\end{align}

\begin{align}
    \Rightarrow H_{\theta_{0}} \nabla_{\theta_{0}} \Delta\Lapl_{(\Set{G}\setminus \Delta \Set{G})} = diag(D_{1}^{-1}E_{1},D_{2}^{-1}E_{2},...,D_{C}^{-1}E_{C})
\end{align}
\begin{align}
    \Rightarrow \hat{W} - {W}_{0} = (w_{1},  w_{2},..,  w_{C}), \quad w_{i} = D_{i}^{-1}E_{i}
\end{align}
\begin{gather}
    D_{j} = \sum_{z_{i} \in \Set{G}} p_{M_i,j}(1-p_{M_i,j})h_{M_i}^Th_{M_i},  \quad
    \quad E_{j} = E_{j}^{(rm)} + E_{j}^{(nei)} \nonumber \nonumber\\
    E_{j}^{(rm)} = \sum\limits_{z_{i} \in \Set{V}^{rm}} \hat{p}_{M_{i},C_{j}}h_{M_{i}}^{T},\\
    E_{j}^{(nei)} = \sum\limits_{z_{i} \in N_{K}(V^{rm}} ( {p}^{\prime}_{M_{i},C_{j}}h_{M_{i}}^{T} - \hat{p}^{\prime}_{M_{i},C_{j}}\hat{h}_{M_{i}}^{T} )
\end{gather}

\section{Hyperparameter Analysis}\label{apd:ablation}
\begin{figure}[h]
\centering
\includegraphics[width=0.45\linewidth]{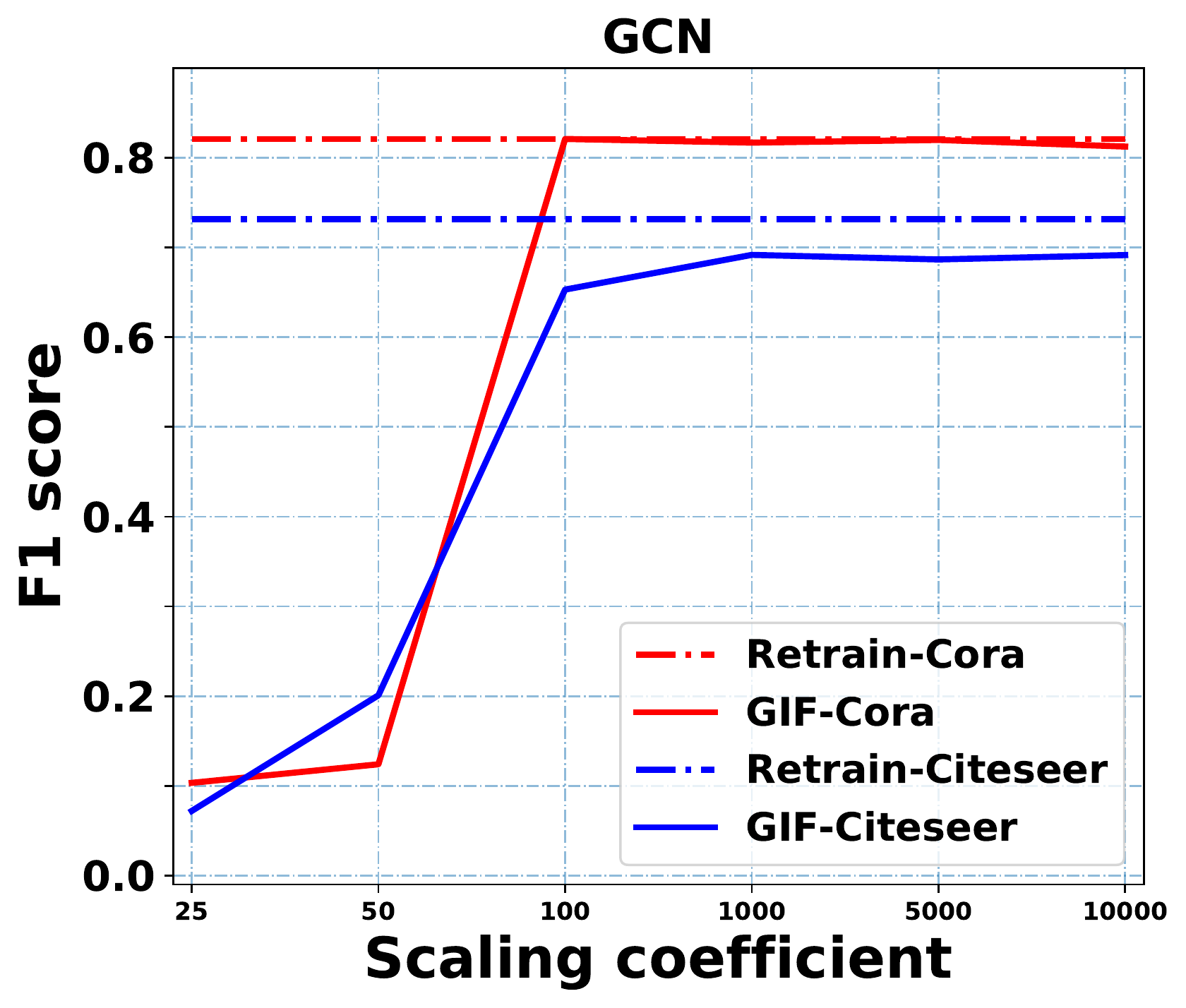}
% \hspace{1in}
\includegraphics[width=0.45\linewidth]{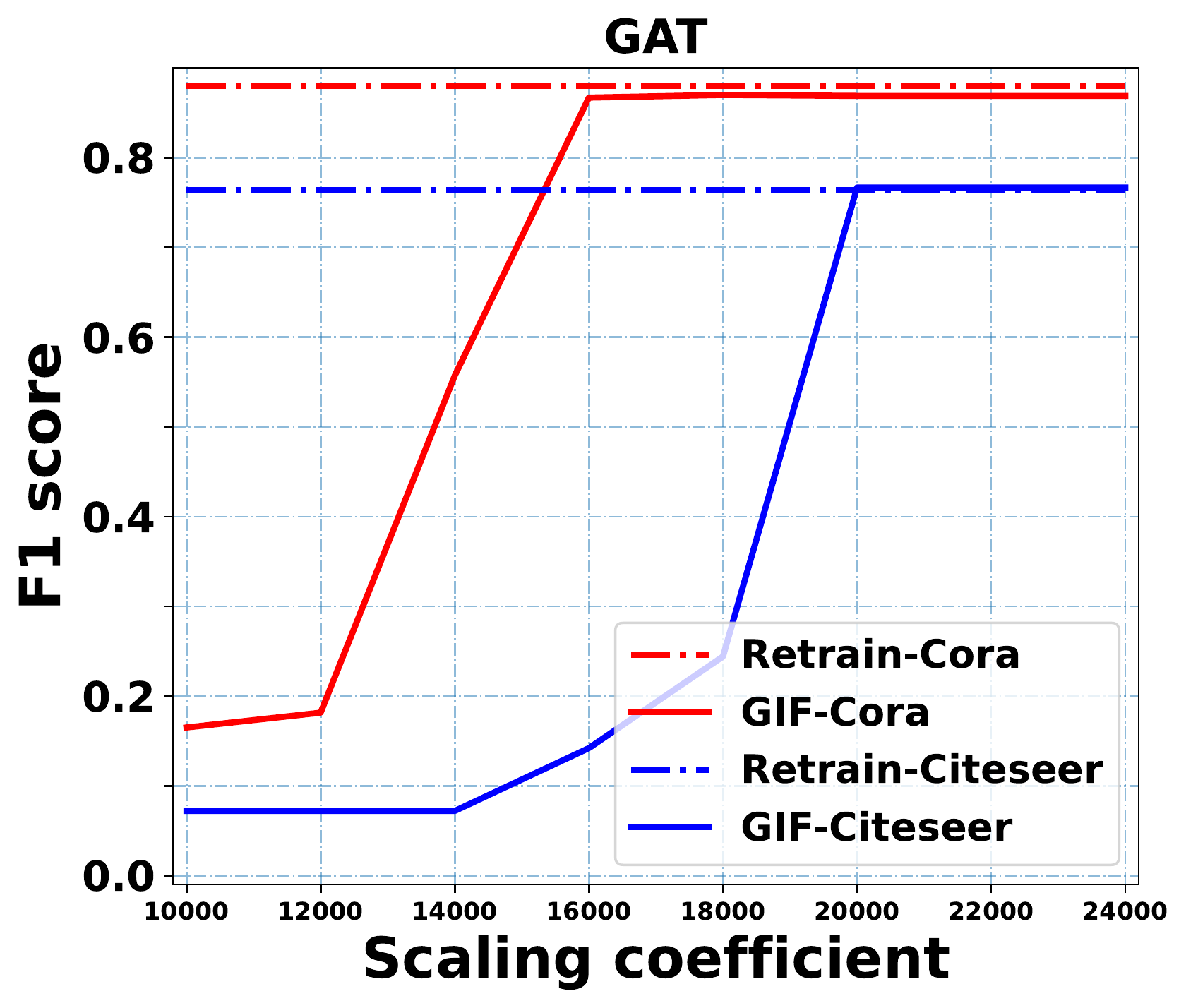}

\caption{Comparsion of F1 scores over different scaling coefficients in GCN and GAT models}
\Description[Impact of scaling coefficient]{Using GCN and GAT as backbone models.}
\label{fig:abl_lambda}
% \vspace{-4mm}
\end{figure}

Since the scaling coefficient $\lambda$ controls the convergence condition of the efficient estimation algorithm, we here investigate how the choice of $\lambda$ influences the utility of the unlearned model in the edge unlearning tasks. We omit other tasks as they exhibit a similar trend.
We fix the unlearning ratio to 0.05 and the iteration number to 100. We compare the F1 score of the retrain version and the unlearned version of GCN and GAT on Cora and Citeseer datasets. As the results shown in Figure~\ref{fig:abl_lambda}, we have the following \textbf{Obs}ervations:
\begin{itemize}[leftmargin=*]
    \item \textbf{Obs5: The scaling coefficient exerts large impacts on the performance of GIF.} Specifically, when $\lambda$ is small, the performance of the unlearned model is poor, which suggests that the convergence condition is not achieved. However, when $\lambda$ exceeds a threshold, the performance of the unlearned model rises rapidly as $\lambda$ increases.
    Thereafter, the performance reaches saturation, that is, increasing $\lambda$ will not bring performance gain. The saturation performance is close to "Retrain", verifying the superiority of GIF.
    In addition, the saturation point of $\lambda$ varies across GNN models and datasets. For example, when implemented on GAT, the peak performance is achieved when $lambda$ is larger than 16000 and 20000 on Cora and Citeseer, respectively; While on GCN, $\lambda=1000$ is sufficient for both Cora and Citeseer. In general, we suggest picking a large $\lambda$ for convergence.
\end{itemize}
% Compared to the hyperparameter $\lambda$, we find that the number of iterations $T$ has less impact on the model performance. We fix $T$ to 100, which is enough for the experiment result to stabilise.

% We study the influence of scaling coefficient $\lambda$ on different datasets and GNN models.
% Figure~\ref{fig:5} shows the experimental results of GIF and Retrain on Cora and Citeseer datasets using the default iteration number of 100.
% We find that a small scaling coefficient $\lambda$ (e.g. smaller than 100 for GCN models) typically leads to unsatisfactory performance. However, when $\lambda$ is chosen to be large enough(e.g. larger than for GAT models), the scores always remain in the optimal range. It is worth noting for different GNN models, the threshold for the scaling coefficient varies(e.g., approximately for 100 GCN models, approximately 20000 for GAT models). As a result, we recommend fixing the number of iterations to 100 and tuning the scaling coefficient $\lambda$ in the range of $\{ 10^{2}, 10^{3}, 10^{4}, 10^{5},\cdots\}$ during the experiment.

\end{document}